\documentclass{article}
\usepackage{amssymb}

\usepackage{latexsym}
\usepackage{multirow}
\usepackage[boxed]{algorithm}
\usepackage{algorithmicx}
\usepackage{algpseudocode}
\usepackage{graphics}
\usepackage{color} 
\usepackage{epsfig} 

\graphicspath{{./FIGS/}}

\newenvironment{pf}[1][Proof]{
\begin{trivlist}
\item[\hskip \labelsep {\bfseries #1}]}{
\end{trivlist}
}

\def\mini{\mathop{\mbox{minimize}}}
\def\maxi{\mathop{\mbox{maximize}}}
\def\IR{\mathbb{R}}

\def\AA{\mathcal{A}}
\def\BB{\mathcal{B}}
\def\XX{\mathcal{X}}
\def\YY{\mathcal{Y}}
\def\ZZ{\mathcal{Z}}
\def\CC{\mathcal{C}}

\def\trace{\mbox{tr}}
\def\rank{\mbox{rank}}

\def\kNN{k\mbox{NN}}

\begin{document}

\title{Classification with Repulsion Tensors:
A Case Study on Face Recognition}

\author{Hawren Fang%
\thanks{Email: \texttt{hrfang@yahoo.com}}
}

\maketitle

\begin{abstract}
We consider dimensionality reduction methods
for face recognition in a supervised setting,
using an image-as-matrix representation.
A common procedure is to project image matrices into a smaller space
in which the recognition is performed.
These methods are often called ``two-dimensional'' in the literature and
there exist  counterparts that use an image-as-vector representation.
When two face images are close to each other in the input space they
may remain close after projection - but this is not desirable in the
situation when  these two images are from different classes, and this 
often affects the recognition performance.
We extend a previously developed `repulsion Laplacean' 
technique based on adding terms to the objective function with the
goal or creation a repulsion energy between 
such images in the projected space.
This scheme, which relies on a repulsion graph,
is generic and can be incorporated into
various two-dimensional methods.
It can be regarded as a multilinear generalization of
the repulsion strategy by Kokiopoulou and Saad
[Pattern Recog., 42 (2009), pp. 2392--2402].
Experimental results demonstrate that
the proposed methodology offers significant recognition improvement
relative to  the underlying  two-dimensional methods.

\end{abstract}

%


\section{Introduction}
\label{sec:intro}

Numerous linear dimensionality reduction methods
have been developed for classification tasks.
Among these are a class  methods which convert the data items to
vectors in numerical form.
For example, in face recognition,
a face image is transformed into
a column vector by concatenating rows or columns.
A linear projector is obtained using the training data.
The projector projects both training and test data
into a lower dimensional space,
in which the recognition task is performed.
Typical examples of these include 
principal component analysis (PCA) \cite{tp:eigenfaces91a,tp:eigenfaces91b},
linear discriminant analysis (LDA) \cite{bhk:fisher97,fisher:lda36,mk:pca_lda01},  
locality preserving projections (LPP) \cite{hn:lpp03,hyunz:laplacian05}, and
neighborhood preserving projections (NPP) \cite{hcyz:npe05}.
There are counterparts of LPP and NPP which enforce
orthogonality constraint on the projector.
They are denoted by OLPP and ONPP \cite{ks:onpp05,ks:onpp07}, respectively.

Two-dimensional projection methods,
treating each image as a matrix, with 
a goal of exploiting spatial redundancy.
In the literature, such methods often utilize  the key word
`two-dimensional' to charaterize them.
They can be categorized into two different types.
The first type of methods aim to
preserve the distinct features of the data matrices
in the projection
\cite{kwtlwv:g2dpca05,ss:hooi07,xyzlzu:csa08,yzfy:2dpca04,ye:glram05}.
These methods can be regarded as  two-dimensional generalizations of
the classical principal component analysis.

The other type of methods
aim to preserve the closeness of neighboring data items
in the projected space.
In a supervised setting,
two items are regarded as neighbors if they share the same class label.
Examples of such methods include
two-dimensional linear discriminant analysis (2D-LDA) \cite{jwz:fisher2d06,ly:fisher2d04},
two-dimensional neighborhood preserving projections (2D-NPP) \cite{ld:2dnpp07,rd:2donpp08}, and
two-dimensional locality preserving projections (2D-LPP) \cite{czkl:2dlpp07,hcn:tsa05,nysp:laplacian2d08}.

\begin{table*}[htbp]
\center{
\caption{Linear dimensionality reduction methods.\label{tbl:alg_summary}}
\begin{tabular}{clclll} \hline
Data & \multicolumn{1}{c}{Method} & Abbrev. & In Face Recog. & References \\ \hline
\multirow{4}{*}{1D}
  & Principal Component Analysis         & PCA & Eigenfaces     & \cite{jolliffe:pca02,tp:eigenfaces91a,tp:eigenfaces91b} \\
  & Linear Discriminant Analysis          & LDA & Fisherfaces    & \cite{bhk:fisher97,fisher:lda36,mk:pca_lda01} \\
  & Locality Preserving Projections     & LPP & Laplacianfaces & \cite{hn:lpp03,hyunz:laplacian05} \\
  & Neighborhood Preserving Projections & NPP & -              & \cite{hcyz:npe05,ks:onpp05,ks:onpp07}\\
\hline
\multirow{4}{*}{2D}
  & Principal Component Analysis         & 2D-PCA & 2D-Eigenfaces     & \cite{kwtlwv:g2dpca05,ss:hooi07,xyzlzu:csa08,yzfy:2dpca04,ye:glram05} \\
  & Linear Discriminant Analysis          & 2D-LDA & 2D-Fisherfaces    & \cite{jwz:fisher2d06,ly:2dlda05,yjl:2dlda03} \\
  & Locality Preserving Projections     & 2D-LPP & 2D-Laplacianfaces & \cite{czkl:2dlpp07,hcn:tsa05,nysp:laplacian2d08} \\
  & Neighborhood Preserving Projections & 2D-NPP & -                 & \cite{rd:2donpp08,ld:2dnpp07} \\
\hline
\end{tabular}
}
\end{table*}

The aforementioned methods are summarized in Table~\ref{tbl:alg_summary},
where the synonyms in the application of face recognition, if any,
are also listed.

An enhanced graph-based dimensionality reduction technique,
{\em repulsion Laplaceans} \cite{ks:repulsion09}, was proposed
to improve the classification methods which use image-as-vector
representation.
The objective is
to repel from other data points that are not from the
same class but that 
close to each other in the input high-dimensional space.

In this paper, we develop a generalized methodology
using {\em repulsion tensors}
to improve the two-dimensional projection methods.
This method is generic and can be applied to various methods,
such as 2D-OLPP, 2D-ONPP, and 2D-LDA.
The experiments on face recognition
demonstrate significant improvements
in the recognition performance
over the existing two-dimensional methods.
Compared to the peers using the image-as-vector representation,
the proposed technique achieves substantial computational savings.

The rest of this paper is organized as follows.
Section~\ref{sec:2dmethods} formulates
the two-dimensional projection methods in tensor form,
establishes the connections between them,
and gives a unified view.
The enhancement with repulsion tensors is presented
in Section~\ref{sec:trepulsion}, which also
gives a unified framework for computing the projectors.
Section~\ref{sec:exp} reports on
experimental results with face recognition. The paper ends with
conclusing remarks and a discussion of future work in
 Section~\ref{sec:end}.
Appendix~\ref{sec:tensor_intro} introduces
the concept of tensors and some related operations
useful in this paper.
Appendix~\ref{sec:1dmethods} reviews
the linear dimensionality reduction methods
using image-as-vector representation.

The notational conventions in this paper are as follows.
We denote scalars and vectors by lowercase letters (e.g., $m,n$),
matrices by uppercase letters (e.g., $U,V$),
and tensors by calligraphic letters (e.g., $\AA,\BB$).
An exception is that $I,J,K$ are used for upper bounds of indices.
For a third order tensor $\AA\in\IR^{I\times J\times K}$,
its $(i,j,k)$ entry is denoted by $a_{ijk}$.
We also use MATLAB-like notation.
For example, $\AA(i,j,k)=a_{ijk}$,
and $\AA(:,:,k)$ is the matrix formed
by elements $a_{ijk}$ for $i=1,\dots,I$ and $j=1,\dots,J$
and the specified $k$.
We use $I_n$ to denote the identity matrix of size $n$-by-$n$,
and $e_n$ to denote the vector of ones of length $n$.
The norm $\|\cdot\|$ indicates the Frobenius norm.

\section{Tensor formulation of 2D methods}
\label{sec:2dmethods}
Table~\ref{tbl:alg_summary} summarizes various known two-dimen\-sional methods
for image subspace analysis.
These methods treat each image as a matrix $X_k\in R^{m_1\times m_2}$.
The general form of the {\em bilateral projection} is
\begin{equation}
\label{eqn:2d_proj}
Y_k = U^T X_kV\in R^{d_1\times d_2},
\quad
k=1,\dots,n,
\end{equation}
where $d_1\leq m_1$ and $d_2\leq m_2$.
The transformation matrices $U$ and $V$ from the training data
are applied to a test image $X$ to obtain a projected image $Y=U^TXV$,
and the recognition is performed by comparing 
$Y$ to $Y_1,\dots,Y_n$.
When $U$ is an identity matrix  ($d_1=m_1$) or 
$V$ is an identity matrix  ($d_2=m_2$),
the projection is said to be  {\em unilateral}.

In this section, we formulate these methods in tensor form.
Readers are referred to \cite{bk:tensor06} for tensor concepts and
operations.  A brief introduction of the main ideas is also
given in Appendix~\ref{sec:tensor_intro}.

Aggregating the matrices in (\ref{eqn:2d_proj}),
we obtain the third order tensors $\XX\in\IR^{m_1\times m_2\times n}$ and
$\YY\in\IR^{d_1\times d_2\times n}$, such that
$\XX(:,:,k)=X_k$ and $\YY(:,:,k)=Y_k$ for $k=1,\dots,n$.
The relation between $\XX$ and $\YY$ can be written succinctly as
\begin{equation}
\label{eqn:tensor_proj}
\YY=\XX \times_1 U^T \times_2 V^T.
\end{equation}

Constraints should be imposed to the transformation matrices $U$ and $V$.
A popular choice is the orthogonality
\begin{equation}
\label{eqn:orth_proj}
U^TU=I_{d_1},\quad V^TV=I_{d_2}.
\end{equation}

To facilitate the  discussion,
we introduce the concept of {\em tensor trace}.
Given a fourth order tensor $\BB=[b_{ijkh}]\in\IR^{I\times J\times I\times J}$,
we define the tensor trace by
\begin{equation}
\label{eqn:tensor_trace}
\trace_{[1,2;3,4]}(\BB)=\sum_{i=1}^I\sum_{j=1}^J b_{ijij},
\end{equation}
where the subscript $[1,2;3,4]$ specifies
the order of the dimensions respect to which the trace is taken.
This is a generalization of the trace of a square matrix.
In this paper we always use the order $[1,2;3,4]$ in the tensor trace.
Hence we abbreviate $\trace_{[1,2;3,4]}(\BB)$ as $\trace(\BB)$.
A useful property which parallels  the relation $\|A\|^2=\trace(AA^T)$ is
$$
\|\AA\|^2=\trace(\langle \AA,\AA\rangle_{[3;3]}),
$$
where $\AA=\IR^{I\times J\times K}$ is a third order tensor, and
$\langle \AA,\AA\rangle_{[3;3]}$ is the mode-$[3;3]$ contracted tensor product.
In the following discussion,
we will use tensor and matrix notation interchangeably.

\subsection{Two-dimensional PCA}
\label{sec:2dpca}

We consider a class of tensor methods which can be regarded
as high order generalizations of the classical PCA
\cite{kwtlwv:g2dpca05,ss:hooi07,xyzlzu:csa08,yzfy:2dpca04,ye:glram05}.
The discussion starts with a bilateral projection method,
also referred to as 
the generalized low rank approximation of matrices (GLRAM) \cite{ye:glram05} or
the concurrent subspaces analysis (CSA) \cite{xyzlzu:csa08}.
This is a special case of
the high order orthogonal iteration (HOOI) of tensors
\cite{lmv:tensor_approx00,ss:hooi07}.

The goal here is to find a reduced representation
$Y_k\in\IR^{d_1\times d_2}$ of each $X_k\in \IR^{m_1\times m_2}$
for $k=1,\dots,n$.
The reconstructed data,
associated with two orthogonal  matrices $U$ and $V$
($U^TU=I_{d_1}$ and $V^TV=I_{d_2}$), is $\widehat{X}_k=UY_kV^T$.
Each $\widehat{X}$ is a rank-$d$ approximation of $X$,
where $d=\min\{d_1,d_2\}$.
The reconstruction errors are measured by
\begin{equation}
\label{eqn:tensor_approx_obj}
\sum_{k=1}^{n}\|X_k-\widehat{X}_k\|^2
=
\sum_{k=1}^{n}\|X_k-UY_kV^T\|^2
\end{equation}
Now let $f_k(Y) = \|X_k-UYV^T\|^2$, which equals
$$
\trace(X_kX_k^T) + \trace(YY^T) -2\,\trace(YU^TX_kV).
$$
Since $f_k(Y)$ is convex,
the minimum of $f_k(Y)$ is achieved by setting $\nabla f_k=0$,
from which we obtain $Y=U^TX_kV$.
Therefore, when (\ref{eqn:tensor_approx_obj}) is minimized,
(\ref{eqn:2d_proj}) is satisfied.

Substituting (\ref{eqn:2d_proj}) into (\ref{eqn:tensor_approx_obj}),
we obtain
$$
\sum_{k=1}^{n}\|X_k-UU^TX_kVV^T\|^2
=
\sum_{k=1}^n\|X_k\|^2-\|U^TX_kV\|^2
=
\|\XX\|^2-\|\XX\times_1 U^T\times_2 V^T\|^2.
$$
As a result, minimizing $\|\XX-\YY\times_1 U\times_2 V\|^2$
is equivalent to maximizing $\|\XX\times_1 U^T\times_2 V^T\|^2$.
We end-up with the following problem: 
\begin{equation}
\label{eqn:tensor_approx_prog}
\left\{
    \begin{array}{ll} \displaystyle
    \maxi_{U,\,V} & \|\XX\times_1 U^T\times_2 V^T\|^2 \\
    \textup{subject to}   & U^TU=I_{d_1},\; V^TV=I_{d_2}.   
    \end{array}
\right.
\end{equation}
Note that we can write $\|\XX\times_1 U^T\times_2 V^T\|^2$
in tensor trace form as
\begin{equation}
\label{eqn:tensor_approx_objx}
\trace(\langle\XX\times_1 U^T\times_2 V^T,\XX\times_1 U^T\times_2 V^T\rangle_{[3;3]}).
\end{equation}

A variant of the above method is called
the generalized two-dimensional PCA \cite{kwtlwv:g2dpca05}.
The formulation is essentially the same as (\ref{eqn:tensor_approx_prog}),
except that before dimensionality reduction,
the data matrices are rigidly translated
so that its  centroid is at the origin.
To be specific,
the translated matrices are
$\widetilde{X}_k=X_k-\bar{X}$ for $i=1,\dots,n$,
where $\bar{X}=\frac{1}{n}\sum_{i=1}^n X_i$ 
is the centroid matrix.
Let $\widetilde{\XX}$ be the translated tensor such that
$\widetilde{\XX}(:,:,k)=\widetilde{X}_k$ for $k=1,\dots,n$.
Then
$$
\widetilde{\XX}
=
\XX-\XX\times_3 \frac{e_n^T}{n} \times_3 e_n
=
\XX-\XX\times_3 (\frac{e_ne_n^T}{n})
=
\XX\times_3 (I_n-\frac{1}{n}e_ne_n^T),
$$
where $e_n\in\IR^n$ is the column vector of ones.
In a similar discussion leading to (\ref{eqn:tensor_approx_prog}),
we obtain the program
\begin{equation}
\label{eqn:tpca_prog}
\left\{
    \begin{array}{ll} \displaystyle
    \maxi_{U,V} & \|\XX\times_1 U^T\times_2 V^T\times_3 J_n\|^2 \\
    \textup{subject to}   & U^TU=I_{d_1},\; V^TV=I_{d_2},
    \end{array}
\right.
\end{equation}
where $J_n=I_n-\frac{1}{n}e_ne_n^T$ is the centering matrix.
The objective function can be written in the form of tensor trace as
\begin{equation}
\label{eqn:tpca_objx}
\trace(\XX\times_1 U^T\times_2 V^T\times_3 J_n,\XX\times_1 U^T\times_2 V^T).
\end{equation}
Note that $J_n$ is a projection matrix.
Hence $J_n^2=J_n=J_n^T$.
This property has been used in deriving (\ref{eqn:tpca_objx}).
We call the dimensionality reduction method which solves (\ref{eqn:tpca_prog})
the 2D-PCA hereafter.

There is no closed form of solution to (\ref{eqn:tensor_approx_prog}) and (\ref{eqn:tpca_prog}).
One can use
the high order orthogonal iteration (HOOI) \cite{lmv:tensor_approx00}
to compute $U$ and $V$.

One may perform 2D-PCA with the projection applied to
only one side of the input image matrices \cite{yzfy:2dpca04}.
For example, we set $V$ as the identity $I_{m_2}$ and therefore
the projection is in the form $Y_i=U^T X_i\in\IR^{d_1\times m_2}$ for
$i=1,\dots,n$.
This unilateral projection is equivalent to
performing PCA on the columns of all images.

\subsubsection{Connection to PCA}

Recall that PCA maximizes the trace of covariance matrix
of the embedded data subject to orthogonal projection.
See, e.g., Appendix~\ref{sec:pca}.
Now we draw the corresponding property of 2D-PCA.   
Substituting (\ref{eqn:tensor_proj}) into
the objective function of (\ref{eqn:tpca_prog}),
we obtain
$$
\|\YY\times_3 J_n\|^2
=
\sum_{k=1}^n \|Y_k-\bar{Y}\|^2
=
\trace\left[\sum_{k=1}^n (Y_k-\bar{Y})(Y_k-\bar{Y})^T\right],
$$
where $\bar{Y}=\sum_{k=1}^n Y_k$. Here
$$
\frac{1}{n}\sum_{k=1}^n (Y_k-\bar{Y})(Y_k-\bar{Y})^T
$$
is called the {\em image covariance matrix},
whose trace is maximized by 2D-PCA
subject to the orthogonality constraints (\ref{eqn:orth_proj}).
If $m_2=1$, then $Y_1,\dots,Y_n$ are indeed column vectors,
in which case 2D-PCA is equivalent to PCA.

\subsection{Two-dimensional LPP}
\label{sec:2dlpp}

LPP and OLPP, reviewed in Appendix~\ref{sec:lpp}, are
linear dimensionality reduction methods
using image-as-vector representation.
Their two-dimensional counterparts,
using image-as-matrix representation
\cite{czkl:2dlpp07,hcn:tsa05,nysp:laplacian2d08}, are presented next.

The step to construct
the symmetric weight matrix $W=[w_{ij}]\in\IR^{n\times n}$
is the same as that in LPP and OLPP.
We also compute the graph Laplacian $L=D-W$,
where $D\in\IR^{n\times n}$ is the diagonal matrix
formed by elements $d_{ii}=\sum_{j=1}^n w_{ij}$ for $i=1,\dots,n$.
See Appendix~\ref{sec:affinity_graph} for details.

LPP and OLPP minimize the objective function (\ref{eqn:lle_obj}) 
in Appendix~\ref{sec:npp}.
Correspondingly, we minimize
\begin{eqnarray}
\label{eqn:2dlpp_obj}
\frac{1}{2}\sum_{i,j=1}^n w_{ij}\|Y_i-Y_j\|^2
& = &
\frac{1}{2}\sum_{i,j=1}^n \trace\left[w_{ij}(Y_i-Y_j)(Y_i-Y_j)^T\right] \\
& = &
\trace\left[\sum_{i=1}^n d_{ii}Y_iY_i^T - \sum_{i,j=1}^n
w_{ij}Y_iY_j^T\right] \nonumber \\
& = &
\trace\left[\sum_{k=1}^{d_2}\sum_{i=1}^n d_{ii}\YY(:,k,i)\YY(:,k,i)^T \right.
\left. -\sum_{k=1}^{d_2}\sum_{i,j=1}^n w_{ij}\YY(:,k,i)\YY(:,k,j)^T\right] \nonumber \\
& = &
\trace\left[\sum_{k=1}^{d_2} \YY(:,k,:)(D-W)\YY(:,k,:)^T\right] \nonumber \\
& = &
\sum_{k=1}^{d_2} \trace\left[\YY(:,k,:)L\YY(:,k,:)^T\right] \nonumber \\
& = &
\trace(\langle\YY\times_3 L,\YY\rangle_{[3;3]}),
\label{eqn:tlpp_objy}
\end{eqnarray}
where the tensor trace in (\ref{eqn:tlpp_objy}) is defined
in (\ref{eqn:tensor_trace}).
Note that the last term (\ref{eqn:tlpp_objy}) is
a tensor generalization of $\trace(YLY^T)$ in (\ref{eqn:lle_obj}).

We can substitute (\ref{eqn:tensor_proj}) into (\ref{eqn:tlpp_objy})
and obtain
\begin{equation}
\label{eqn:tlpp_objx}
\trace(\langle\XX\times_1 U^T\times_2 V^T\times_3 L,\XX\times_1 U^T\times_2 V^T\rangle_{[3;3]}).
\end{equation}

Constraints are required in the minimization of (\ref{eqn:tlpp_objx}).
For example, we can impose the orthogonality
$U^TU=I_{d_1}$ and $V^TV=I_{d_2}$ in (\ref{eqn:orth_proj}).
This option leads to the 2D-OLPP method,
\begin{equation}
\label{eqn:2dolpp_prog}
\left\{
  \begin{array}{ll} \displaystyle
  \mini_{U,V} & \displaystyle \trace(\langle\YY\times_3 L,\YY\rangle_{[3;3]}) \\
  \textup{subject to}   & \YY=\XX\times_1 U\times_2 V, \\
                        & U^TU=I_{d_1},\; V^TV=I_{d_2}.
  \end{array}
\right.
\end{equation}
How to solve this optimization problem (\ref{eqn:2dolpp_prog}) will be discussed in a unified framework in Section~\ref{sec:uframe}.

Alternatively, we may consider concurrently maximizing
\begin{equation}
\label{eqn:tlpp_ydy}
\trace(\langle\XX\times_1 U^T\times_2 V^T\times_3 D,\XX\times_1 U^T\times_2 V^T\rangle_{[3;3]}),
\end{equation}
which corresponds to $\trace(YDY^T)$ in the LPP program (\ref{eqn:lpp_prog}).
Minimizing the ratio of (\ref{eqn:tlpp_objx}) to (\ref{eqn:tlpp_ydy})
subject to $U^TU=I_{d_1}$ and $V^TV=I_{d_2}$
is a tensor generalization of the trace ratio optimization problem
\cite{nbs:trace_ratio10,wyxth:trace_ratio07}.
Rather than solving this challenging problem,
we will give a workaround in a unified framework in Section~\ref{sec:uframe}.
This corresponds to the alternating algorithm in \cite{hcn:tsa05}.
We call the resulting method 2D-LPP.

When $m_2=1$, $X_1,\dots,X_n$ are indeed column vectors, and
the data tensor $\XX$ can be represented
by a matrix $[X_1,\dots,X_n]$, in which case
2D-LPP is reduced to
the formulation of LPP (\ref{eqn:lpp_prog}),
and 2D-OLPP is equivalent to OLPP.

\subsection{Two-dimensional NPP}
\label{sec:2dnpp}

Recall that NPP and ONPP, reviewed in Appendix~\ref{sec:npp}, are
linear dimensionality reduction methods
that use the image-as-vector representation.
Here we present their two-dimensional counterparts,
using the image-as-matrix representation \cite{ld:2dnpp07,rd:2donpp08}.

Firstly, we compute a weight matrix $W=[w_{ij}]\in\IR^{n\times n}$
in the same way as that of NPP and ONPP.
See Appendix~\ref{sec:affinity_graph} for a discussion.
Then we consider the objective function
to minimize
\begin{eqnarray}
\label{eqn:2dnpp_obj}
\sum_{i=1}^n \|Y_i - \sum_{j=1}^n w_{ij}Y_j\|^2
& = &
\sum_{i=1}^n \|\YY(:,:,i) - \sum_{j=1}^n w_{ij}\YY(:,:,j)\|^2 \\
& = &
\sum_{k=1}^{d_2}\sum_{i=1}^n\|\YY(:,k,i) - \sum_{j=1}^n w_{ij}\YY(:,k,j)\|^2 \nonumber \\
& = &
\sum_{k=1}^{d_2}\|\YY(:,k,:) - \YY(:,k,:)\times_3 W\|^2 \nonumber \\
& = &
\sum_{k=1}^{d_2}\|\YY(:,k,:)\times_3(I_n-W)\|^2 \nonumber \\
& = &
\|\YY\times_3(I_n-W)\|^2, \label{eqn:tnpp_obj0}
\end{eqnarray}
which parallels the objective function of NPP and ONPP (\ref{eqn:lle_obj}).

To match the format of (\ref{eqn:tlpp_objy}),
we rewrite (\ref{eqn:tnpp_obj0}) in the tensor trace form as
\begin{equation}
\label{eqn:tnpp_objy}
\trace(\langle\YY\times_3 H,\YY\rangle_{[3;3]}),
\end{equation}
where $H=(I_n-W)^T(I_n-W)$.
We can further substitute (\ref{eqn:tensor_proj}) into (\ref{eqn:tnpp_objy})
and obtain
\begin{equation}
\label{eqn:tnpp_objx}
\trace(\langle\XX\times_1 U^T\times_2 V^T\times_3 H,\XX\times_1 U^T\times_2 V^T\rangle_{[3;3]}).
\end{equation}

Again, one can impose the orthogonality constraints $U^TU=I_{d_1}$ and $V^TV=I_{d_2}$,
yielding the 2D-ONPP method.
Another option is to concurrently maximizing
\begin{equation}
\label{eqn:tnpp_yy0}
\|\XX\times_1 U^T\times_2 V^T\|^2.
\end{equation}
We would like to write (\ref{eqn:tnpp_yy0}) in tensor trace form as
\begin{equation}
\label{eqn:tnpp_yy}
\trace(\langle\XX\times_1 U^T\times_2 V^T,\XX\times_1 U^T\times_2 V^T\rangle_{[3;3]}),
\end{equation}
which is same as the objective function of GLRAM and CSA (\ref{eqn:tensor_approx_objx}).

Minimizing the ratio (\ref{eqn:tnpp_objx}) to (\ref{eqn:tnpp_yy})
subject to $U^TU=I_{d_1}$ and $V^TV=I_{d_2}$ is
a challenging tensor trace ratio optimization problem.
A workaround will be given in the unified framework in Section~\ref{sec:uframe}.
This corresponds to the alternating algorithm in \cite{ld:2dnpp07}.
The resulting method is called 2D-NPP.

Note that if $m_2=1$, then $X_1,\dots,X_n$ are indeed column vectors,
and the data tensor $\XX$ can be represented
by a matrix $[X_1,\dots,X_n]$, in which case
2D-NPP and 2D-ONPP are reduced to
the formulations of NPP (\ref{eqn:npp_prog})
and ONPP (\ref{eqn:onpp_prog}), respectively.

\subsubsection{Connection to 2D-PCA}

Comparing the 2D-ONPP method to the 2D-PCA program (\ref{eqn:tpca_prog}),
there are only two differences.
Firstly, 2D-ONPP `minimizes' (\ref{eqn:tnpp_objx}), whereas
2D-PCA `maximizes' (\ref{eqn:tpca_objx}).
Both impose the orthogonality constraints $U^TU=I_{d_1}$ and $V^TV=I_{d_2}$.
Secondly, the multiplication $\times_3 (I_n-\frac{1}{n}e_ne_n^T)$
in (\ref{eqn:tpca_objx})
corresponds to the multiplication $\times_3 (I_n-W)$ in
(\ref{eqn:tnpp_objx}).
If a complete graph is used and the weights are uniformly
distributed, then $w_{ij}=\frac{1}{n}$ for $i,j=1,\dots,n$,
yielding $W=\frac{1}{n}e_ne_n^T$. In this case,
the objective functions of 2D-PCA and 2D-ONPP are identical.
This is a property similar to the connection between PCA and ONPP
\cite{ks:onpp05,ks:onpp07}.

\subsection{Two-dimensional LDA}
\label{sec:2dlda}

We now present the two-dimensional counterpart of LDA \cite{yjl:2dlda03},
using image-as-matrix representation
instead of image-as-vector representation.

Suppose we are given training data matrices $X_1,\dots,X_n\in\IR^{m_1\times m_2}$.
Each data sample $X_i$ is associated with a class label $c(i)$.
For each class $j=1,\dots,c$,
there is an index set
$$
\CC_j=\{i:c(i)=j\},
$$
whose size is denoted by $n_j=|\CC_j|$.

The reduced data is obtained from the bilateral transformation
$Y_k=U^TX_kV$ for $k=1,\dots,n$.
The mean of each class $j$ is denoted by
$\bar{Y}_j=\frac{1}{n_j}\sum_{i\in\CC_j}Y_i$,
and the global mean is $\bar{Y}=\frac{1}{n}\sum_{i=1}^n Y_i$.
The within-scatter measure is defined by
\begin{equation}
\label{eqn:tlda_Dw0}
D_w=\trace(\sum_{j=1}^c\sum_{i\in\CC_j}(Y_i-\bar{Y}_j)(Y_i-\bar{Y}_j)^T),
\end{equation}
and the between-scatter measure is
\begin{equation}
\label{eqn:tlda_Db0}
D_b=\trace(\sum_{j=1}^c n_j(\bar{Y}-\bar{Y}_j)(\bar{Y}-\bar{Y}_j)^T).
\end{equation}

Note that if $m_2=1$, then $X_1,\dots,X_n$ are indeed column vectors
and the data tensor $\XX$ can be represented
by a matrix $[X_1,\dots,X_n]$, in which case
$D_w$ in (\ref{eqn:tlda_Dw0}) and $D_b$ in (\ref{eqn:tlda_Db0})
are the same as $\trace(U^TS_wU)$ and $\trace(U^TS_bU)$
in LDA, where $S_w$ and $S_b$ are defined in
(\ref{eqn:Sw}) and (\ref{eqn:Sb}), respectively.
Conceptually, the goal is to minimize $D_w$ and maximize $D_b$,
corresponding to minimizing $\trace(U^TS_wU)$ and maximizing
$\trace(U^TS_bU)$ in LDA.

To write (\ref{eqn:tlda_Dw0}) and (\ref{eqn:tlda_Db0})
in the tensor trace form,
we aggregate $Y_1,\dots,Y_n$ to form a data tensor $\YY$
such that $\YY(:,:,k)=Y_k$.
For each class $j$,
we form a data tensor $\YY_j$ to store all $Y_i$ with $c(i)=j$.
In MATLAB-like notation,
$
\YY_j=\YY(:,:,c(i)==j).
$
We also define a corresponding translated data tensor
$\widetilde{\YY}_j$ by
$\widetilde{\YY}_j(:,:,k)=\YY_j(:,:,k)-\bar{Y}_j$
for $k=1,\dots,n_j$.
The relation between $\widetilde{\YY}_j$ and $\YY_j$
can be written as
$$
\widetilde{\YY}_j
=
\YY_j-\YY_j\times_3 \frac{e_{n_j}^T}{n_j} \times_3 e_{n_j}
=
\YY-\YY\times_3 (\frac{e_{n_j}e_{n_j}^T}{n_j})
=
\YY\times_3 (I_{n_j}-\frac{1}{n_j}e_{n_j}e_{n_j}^T).
$$
Hence we have
$$
\trace(\sum_{i\in\CC_j}(Y_i-\bar{Y}_j)(Y_i-\bar{Y}_j)^T)
=
\sum_{i\in\CC_j}\|Y_i-\bar{Y}_j\|^2 = \|\widetilde{\YY}_j\|^2
=
\trace(\langle\widetilde{\YY}_j,\widetilde{\YY}_j\rangle_{[3;3]})
=
\trace(\langle\YY_j\times_3 J_{n_j},\YY_j\rangle_{[3;3]}),
$$
where $J_{n_j}=I_{n_j}-\frac{1}{n_j}e_{n_j}e_{n_j}^T$.
Note that since $J_{n_j}$ is a projection matrix,
$J_{n_j}^T=J_{n_j}=J_{n_j}^2$.
This property has been used for the last equality.

Now we can rewrite (\ref{eqn:tlda_Dw0}) as
\begin{equation}
\label{eqn:tlda_Dwy}
D_w
=
\sum_{j=1}^c \trace(\sum_{i\in\CC_j}(Y_i-\bar{Y}_j)(Y_i-\bar{Y}_j)^T)
=
\sum_{j=1}^c \trace(\langle\YY_j\times_3 J_{n_j},\YY_j\rangle_{[3;3]})
=
\trace(\langle\YY\times_3 S,\YY\rangle_{[3;3]}),
\end{equation}
where in MATLAB-like notation,
$S(\CC_k,\CC_k)=J_{n_k}$ for $k=1,\dots,c$, and
$S(i,j)=0$ if $c(i)\neq c(j)$.
Substituting (\ref{eqn:tensor_proj}) into (\ref{eqn:tlda_Dwy}),
we obtain the tensor trace form of $D_w$ as
\begin{equation}
\label{eqn:tlda_Dwx}
D_w=
\trace(\langle\XX\times_1 U^T\times_2 V^T\times_3 S,\XX\times_1 U^T\times_2V^T\rangle_{[3;3]}).
\end{equation}
The matrix $S$ actually is the Laplacian matrix
of a label graph with even weights \cite{nysp:laplacian2d08}.
To be precise, let $W=I_n-S$ with $I_n\in\IR^{n\times n}$ the identity.
Then $W=[w_{ij}]\in\IR^{n\times n}$ satisfies
\begin{equation}
\label{eqn:lda_weights}
w_{ij}=\left\{
\begin{array}{ll}
1/n_k & \mbox{if }c(i)=c(j); \\
0 & \mbox{otherwise}.
\end{array}
\right.
\end{equation}
This matrix $S$ has rank $n-c$.
See, for example, \cite{hyunz:laplacian05,ks:onpp05,ks:onpp07}.
The discussion gives a connection to 2D-LPP.

Now we consider $D_b$ in (\ref{eqn:tlda_Db0}), where
\begin{eqnarray*}
\sum_{j=1}^c n_j(\bar{Y}-\bar{Y}_j)(\bar{Y}-\bar{Y}_j)^T
& = &
 (\sum_{j=1}^c n_j)\bar{Y}\bar{Y}^T
 -(\sum_{j=1}^c n_j\bar{Y}_j)\bar{Y}^T
 -\bar{Y}(\sum_{j=1}^c n_j\bar{Y}_j)^T
 + \sum_{j=1}^c n_j \bar{Y}_j\bar{Y}_j^T \\
& = &
 - n\bar{Y}\bar{Y}^T
 + \sum_{j=1}^c n_j \bar{Y}_j\bar{Y}_j^T.
\end{eqnarray*}
Hence we can rewrite (\ref{eqn:tlda_Db0}) as
\begin{equation}
\label{eqn:tlda_Dbn}
D_b
=
-n\,\trace(\bar{Y}\bar{Y}^T) + \sum_{j=1}^c n_j \trace(\bar{Y}_j\bar{Y}_j^T)
=
-n\|\bar{Y}\|^2 + \sum_{j=1}^c n_j \|\bar{Y}_j\|^2.
\end{equation}
The two terms in (\ref{eqn:tlda_Dbn}) can be written in tensor trace form 
as follows. First,
$$
n\|\bar{Y}\|^2
=
n\|\YY\times_3 \frac{e^T}{n}\|^2 \\
=
\trace(\langle\YY\times_3 \frac{e_ne_n^T}{n},\YY\rangle_{[3;3]}).
$$
Then we have
\begin{equation}
\label{eqn:tlda_Db1}
\|\YY\|^2-n\|\bar{Y}\|^2
=
\trace(\langle\YY\times_3 J_n,\YY\rangle_{[3;3]}),
\end{equation}
where $J_n=I_n-\frac{e_ne_n^T}{n}$.
Note that (\ref{eqn:tlda_Db1}) is the same as
the objective function (\ref{eqn:tpca_objx}) of 2D-PCA.
Secondly,
$$
\sum_{j=1}^c n_j \|\bar{Y}_j\|^2
=
\sum_{j=1}^c \trace(\langle\YY_j\times_3\frac{e_{n_j}e_{n_j}^T}{n_j}),\YY\rangle_{[3;3]})
=
\trace(\langle\YY\times_3 W,\YY\rangle_{[3;3]}),
$$
where $W=[w_{ij}]\in\IR^{n\times n}$ is defined in (\ref{eqn:lda_weights}).
Then we have
\begin{equation}
\label{eqn:tlda_Db2}
\|\YY\|^2-\sum_{j=1}^c n_j \|\bar{Y}_j\|^2
=
\trace(\langle\YY\times_3 S,\YY\rangle_{[3;3]}),
\end{equation}
where $S=I_n-W$.
Note that (\ref{eqn:tlda_Db2}) is identical to (\ref{eqn:tlda_Dwy}).
Substituting (\ref{eqn:tlda_Db1}) and (\ref{eqn:tlda_Db2}) into (\ref{eqn:tlda_Dbn}),
we obtain
\begin{equation}
\label{eqn:tlda_Dby}
D_b=\trace(\langle\YY\times_3 (J_n-S),\YY\rangle_{[3;3]}).
\end{equation}
Substituting (\ref{eqn:tensor_proj}) into (\ref{eqn:tlda_Dby}), we obtain
the tensor trace form of $D_b$ as
\begin{equation}
\label{eqn:tlda_Dbx}
D_b=
\trace(\langle\XX\times_1 U^T\times_2 V^T\times_3 (J_n-S),\XX\times_1 U^T\times_2V^T\rangle_{[3;3]}).
\end{equation}

To obtain $U$ and $V$,
one could maximize the ratio of (\ref{eqn:tlda_Dwx}) to (\ref{eqn:tlda_Dbx})
subject to $U^TU=I_{d_1}$ and $V^TV=I_{d_2}$,
where $Y_k=U^TX_kV$ for $k=1,\dots,n$.
Instead of solving this challenging optimization problem,
we will solve a related problem in a unified framework in Section~\ref{sec:uframe}.
This corresponds to the alternating algorithm in \cite{yjl:2dlda03}.

\subsection{An unified view}

All the methods discussed in this section
can be categorized into three cases.
\begin{enumerate}
\item
The first case is to minimize
$\trace(\langle\YY\times_3 A,\YY\rangle_{[3;3]})$.
The methods are 2D-OLPP and 2D-ONPP.
\item
The second case is to maximize
$\trace(\langle\YY\times_3 B,\YY\rangle_{[3;3]})$.
The methods are GLRAM/CSA and 2D-PCA.
\item
The last case is to minimize
$\trace(\langle\YY\times_3 A,\YY\rangle_{[3;3]})$
and to maximize
$\trace(\langle\YY\times_3 B,\YY\rangle_{[3;3]})$
concurrently.
The methods are 2D-LPP, 2D-NPP, and 2D-LDA.
\end{enumerate}

\begin{table*}[htbp]
\center{
\caption{Minimization of $\trace(\langle\YY\times_3 A,\YY\rangle_{[3;3]})$ and
maximization of $\trace(\langle\YY\times_3 B,\YY\rangle_{[3;3]})$.\label{tbl:2dmethods_summary}}
\begin{tabular}{l|cc|cc} \hline
 \multicolumn{1}{c|}{Method} & \makebox[1.825in]{Matrix $A$} & Ref. & \makebox[1.825in]{Matrix $B$} & Ref. \\ \hline
  GLRAM/CSA  & - & & $I_n$ & (\ref{eqn:tensor_approx_objx}) \\
  2D-PCA     & - & & $J_n=I_n-\frac{1}{n}e_ne_n^T$ & (\ref{eqn:tpca_objx}) \\
  2D-OLPP    & $L=D-W$ & (\ref{eqn:tlpp_objx}) & - & \\
  2D-LPP     & $L=D-W$ & (\ref{eqn:tlpp_objx}) & $D$ & (\ref{eqn:tlpp_ydy}) \\
  2D-ONPP    & $H=(I_n-W)^T(I_n-W)$ & (\ref{eqn:tnpp_objx}) & - & \\
  2D-NPP     & $H=(I_n-W)^T(I_n-W)$ & (\ref{eqn:tnpp_objx}) & $I_n$ & (\ref{eqn:tnpp_yy}) \\
  2D-LDA     & $S=I_n-W$ & (\ref{eqn:tlda_Dwx}) & $J_n-S$ & (\ref{eqn:tlda_Dbx}) \\
\hline
  2D-OLPP-R  & $L-\beta L^{(r)}$ & & - &  \\
  2D-LPP-R   & $L-\beta L^{(r)}$ & & $D$ &  \\
  2D-ONPP-R  & $H-\beta L^{(r)}$ & & - &  \\
  2D-NPP-R   & $H-\beta L^{(r)}$ & & $I_n$ &  \\
  2D-LDA-R   & $S-\beta L^{(r)}$ & & $J_n-S$ &  \\
\hline
\end{tabular}
}
\end{table*}

Table~\ref{tbl:2dmethods_summary} lists the corresponding
matrices $A$ and $B$ and their references of all these methods.
Each `-' indicates that there is no matrix $A$ or no matrix $B$,
implying the single objective of maximization or minimization.
The methods using repulsion tensors,
to be described in the next section, are also listed.
The next section will also give a unified framework
for computing the projectors $U$ and $V$
of all these methods.

\section{Enhancement by repulsion tensors}
\label{sec:trepulsion}

Classical linear dimensionality reduction methods, such as LDA and LPP,
project the data in vector form into a lower dimensional space.
There are cases that two between-class data points
which are close each other in the high dimensional space
remain close after being projected to the lower dimensional space.
An enhanced dimensionality reduction technique,
called the repulsion Laplaceans \cite{ks:repulsion09},
was proposed to address such issues.
Here we develop a corresponding two-dimensional method
for classification of data matrices.

\subsection{Repulsion graph}
\label{sec:repulsion}

The methodology presented here is based on the use of repulsion graphs
\cite{ks:repulsion09}.
The technique requires two graphs that define the repulsion graph.
\begin{enumerate}
\item
The first
is the supervised label graph $G=(V,E)$
such that $(i,j)\in E$
if data items $i$ and $j$ have the same label.
\item
The second is an unsupervised affinity graph $G^{(a)}=(V,E^{(a)})$
such that $(i,j)\in E^{(a)}$ if data items $i$ and $j$
are close to each other in the input space.
In practice, we construct a $\kNN$ graph for $G^{(a)}$.
\item
The repulsion graph is 
denoted by $G^{(r)}=(V,E^{(r)})$, and its edges are defined by
$(i,j)\in E^{(r)}$ if and only if $(i,j)\in E^{(a)}$ and  $(i,j)\notin E$.
\end{enumerate}

Note that all $G$, $G^{(a)}$ and $G^{(r)}$ are undirected and
share the same vertices $V=\{1,\dots,n\}$ as data indices.
The high level idea is that if 
 items $i$ and $j$ are not in the same class 
($(i,j)\notin E$) but are close to each other ($(i,j)\in E^{(a)}$) 
we should add a penalty force to `repel' them from
each other in the projection process.

We will eventually construct a graph Laplacian $L^{(r)}$
of the repulsion graph $G^{(r)}=(V,E^{(r)})$.
Hence each edge $(i,j)\in E^{(r)}$ is assigned with a weight
$w_{ij}^{(r)}>0$.
In practice, we use the Gaussian weights (\ref{eqn:gaussian_weights});
see Appendix~\ref{sec:affinity_graph}.
We also let $w_{ij}^{(r)}=0$ if $(i,j)\notin E^{(r)}$ .
The weights form a weight matrix
$W^{(r)}=[w_{ij}^{(r)}]\in\IR^{n\times n}$.
The repulsion Laplacian matrix is
$L^{(r)}=D^{(r)}-W^{(r)}$, where $D^{(r)}=W^{(r)}e_n$.

This matrix $L^{(r)}$ has been utilized to
modify linear dimensionality reduction methods
to project vector data \cite{ks:repulsion09}.
For example, given the training data $X=[x_1,\dots,x_n]\in\IR^{m\times n}$,
the OLPP method minimizes $\trace(U^TXLX^TU)$ subject to $U^TU=I_d$ ($d<m$),
where $L\in\IR^{n\times n}$ is the graph Laplacian.
The improvement is achieved by incorporating $L^{(r)}$
into the objective function.
To be precise, we minimize $\trace(U^TX(L-\beta L^{(r)})X^TU)$
subject to $U^TU=I_d$, where $\beta>0$ is a preset penalty parameter.
The enhanced method is denoted by OLPP-R.
This technique can also be used to improve LDA and ONPP in the same way;
see \cite{ks:repulsion09} for details.

\subsection{Repulsion tensor}

Recall that in the common framework of the two-dimensional methods,
we project data matrix $X_k\in\IR^{m_1\times m_2}$
to $Y_k\in\IR^{d_1\times d_2}$ via $Y_k=U^TX_kV$
for $k=1,\dots,n$.
Except for GLRAM/CSA and 2D-PCA,
the methods presented in Section~\ref{sec:2dmethods} all
have an objective function to minimize in the form
\begin{equation}
\label{eqn:tensor_obja}
\trace(\langle\XX\times_1 U\times_2 V\times_3 A,\XX\times_1 U\times_2 V\rangle_{[3;3]}).
\end{equation}

Following the discussion in Section~\ref{sec:repulsion},
we have a repulsion graph $G^{(r)}=(V^{(r)},E^{(r)})$
and its Laplacian $L^{(r)}=D^{(r)}-W^{(r)}\in\IR^{n\times n}$.
Recall that the goal in 2D-LPP and 2D-OLPP,
is to make neighboring points defined by $G=(V,E)$
close to each other in the projected space. 
Therefore we minimize (\ref{eqn:2dlpp_obj}).
Here the objective is to repel the neighboring points
defined by $G^{(r)}=(V^{(r)},E^{(r)})$.
Hence we maximize
\begin{equation}
\label{eqn:2drepulsion_obj}
\frac{1}{2} \sum_{i=1}^n w_{ij}^{(r)}\|Y_i - Y_j\|^2
=
\trace(\langle\YY\times_3 L^{(r)},\YY\rangle_{[3;3]}).
\end{equation}
The equation (\ref{eqn:2drepulsion_obj}) can be seen from the derivation of (\ref{eqn:tlpp_objy}).
Substituting (\ref{eqn:tensor_proj}) into (\ref{eqn:2drepulsion_obj}),
we obtain
\begin{equation}
\label{eqn:trepulsion_objx}
\trace(\langle\XX\times_1 U^T\times_2 V^T \times_3 L^{(r)},\XX\times_1 U^T\times_2V^T\rangle_{[3;3]}),
\end{equation}
where
$
\langle\XX\times_1 U^T\times_2 V^T \times_3 L^{(r)},\XX\times_1 U^T\times_2 V^T\rangle_{[3;3]}
$
is called the {\em repulsion tensor}.

As an illustration, we incorporate the maximization of (\ref{eqn:trepulsion_objx})
into the minimization of (\ref{eqn:tlpp_objx}).
Subtracting $\beta\cdot$(\ref{eqn:trepulsion_objx}) from
(\ref{eqn:tlpp_objx}), we obtain
\begin{equation}
\label{eqn:tlppr_objx}
\trace(\langle\XX\times_1 U^T\times_2 V^T \times_3 (L-\beta L^{(r)}),\XX\times_1 U^T\times_2 V^T\rangle_{[3;3]}),
\end{equation}
which is the objective function to minimize.
Here the penalty parameter $\beta>0$ is preset.
If we impose the orthogonality constraints $U^TU=I_{d_1}$ and $V^TV=I_{d_2}$,
then the resulting method is called 2D-OLPP-R.
If we follow 2D-LPP to maximize (\ref{eqn:tlpp_ydy}) concurrently,
then the method is called 2D-LPP-R.

In 2D-ONPP, 2D-NPP, and 2D-LDA,
there is a matrix which plays the role of graph Laplacian
in 2D-LPP and 2D-OLPP.
Hence the repulsion technique described above can be applied.
We denote the modified methods by
2D-ONPP-R, 2D-NPP-R, and 2D-LDA-R,
which incorporate the repulsion technique.
The column `Matrix $A$' of Table~\ref{tbl:2dmethods_summary}
lists all the matrices in the role of graph Laplacian.

\subsection{The unified framework}
\label{sec:uframe}

All the methods in Table~\ref{tbl:2dmethods_summary} can be 
categorized into three categories.
The first has the objective is to minimize
\begin{equation}
\label{eqn:ttraceA}
\trace(\langle\XX\times_1 U^T\times_2 V^T\times_3 A,\XX\times_1 U^T\times_2 V^T\rangle_{[3;3]})
\end{equation}
subject to $U^TU=I_{d_1}$ and $V^TV=I_{d_2}$.
Methods in this category are
2D-OLPP, 2D-ONPP, 2D-OLPP-R and 2D-ONPP-R.
In the second case,
the objective is to maximize
\begin{equation}
\label{eqn:ttraceB}
\trace(\langle\XX\times_1 U^T\times_2 V^T\times_3 B,\XX\times_1 U^T\times_2 V^T\rangle_{[3;3]})
\end{equation}
subject to $U^TU=I_{d_1}$ and $V^TV=I_{d_2}$.
Methods in this category are
GLRAM/CSA and 2D-PCA.
In the last case,
the objective is to minimize (\ref{eqn:ttraceA}) and maximize (\ref{eqn:ttraceB})
concurrently.
One could also impose the orthogonality constraints
$U^TU=I_{d_1}$ and $V^TV=I_{d_2}$.
However this leads to a challenging optimization problem.
A practical workaround will be discussed later in this section.
Methods in the last category are
2D-LPP, 2D-NPP, 2D-LDA, 2D-LPP-R, 2D-NPP-R, and 2D-LDA-R.

We start with the first case.
There is no closed form solution to the problem of minimizing
(\ref{eqn:ttraceA}) subject to $U^TU=I_{d_1}$ and $V^TV=I_{d_2}$.
An alternating process to solve this problem is as follows.

For simplicity, we fix $U\in\IR^{m_1\times d_1}$ and
minimize (\ref{eqn:ttraceA}) in terms of $V\in\IR^{m_2\times d_2}$.
Letting
$\ZZ_1=\XX\times_1 U^T$
and substituting it into (\ref{eqn:ttraceA}),
we obtain
\begin{eqnarray}
\label{eqn:ttraceA1}
 \trace(\langle\ZZ_1\times_2 V^T\times_3 A,\ZZ_1\times_2 V^T\rangle_{[3;3]})
& = &
\trace(\sum_{i=1}^{d_1}(V^T \ZZ_1(i,;,:) A)(V^T \ZZ_1(i,:,:))^T) \\
& = &
\trace(V^T (\sum_{i=1}^{d_1}\ZZ_1(i,:,:) A\ZZ_1(i,:,:)^T) V). \nonumber
\end{eqnarray}
Hence the minimizer of (\ref{eqn:ttraceA1}) subject to $V^TV=I_{d_2}$
consists of the bottom $d_2$ eigenvectors
corresponding to the smallest eigenvalues of
\begin{equation}
\label{eqn:A1}
A_1=\sum_{i=1}^{d_1}\ZZ_1(i,:,:) A\ZZ_1(i,:,:)^T.
\end{equation}
If we take the top eigenvectors,
we obtain the maximizer of (\ref{eqn:ttraceA1}) subject to $V^TV=I_{d_2}$.

Likewise, we can fix $V\in\IR^{m_2\times d_2}$ and minimize (\ref{eqn:ttraceA}) in terms of
$U\in\IR^{m_1\times d_1}$.
Similarly, we let
$\ZZ_2=\XX\times_2 V^T$,
and the minimizer subject to $U^TU=I_{d_1}$
consists of the bottom $d_1$ eigenvectors of
\begin{equation}
\label{eqn:A2}
A_2=\sum_{j=1}^{d_2}\ZZ_2(:,j,:) A\ZZ_2(:,j,:)^T.
\end{equation}

We can solve the two sub-problems alternatively.
The discussion leads to an alternating process to minimize (\ref{eqn:ttraceA})
subject to $U^TU=I_{d_1}$ and $V^TV=I_{d_2}$.
The pseudo-code is given in Algorithm~\ref{alg:alt1}.
This is the high order orthogonal iteration (HOOI) of tensors
\cite{lmv:tensor_approx00,ss:hooi07}.

\begin{algorithm}[htbp]
\caption{Alternating process \#1 for $U,V$.
\label{alg:alt1}}
\begin{algorithmic}
  \State {\bf input:} $\XX\in\IR^{m_1\times m_2\times n}$,
                      $A\in\IR^{n\times n}$.
  \State {\bf output:} projectors $U\in\IR^{m_1\times d_1}$, $V\in\IR^{m_2\times d_2}$.
  \State Set initial $U\gets I_{m_1}$;
  \Repeat
    \State $\ZZ_1 \gets \XX \times_1 U^T$;
    \State $\displaystyle A_1\gets \sum_{i=1}^{d_1} \ZZ_1(i,:,:)A\ZZ_1(i,:,:)^T$;
    \State Compute the bottom $d_2$ eigenvectors of
$
A_1v_i=\lambda_iv_i;
$
    \State $V\gets [v_1,\dots,v_{d_2}]$;
    \State $\ZZ_2 \gets \XX \times_2 V^T$;   
    \State $\displaystyle A_2\gets \sum_{j=1}^{d_2} \ZZ_2(:,j,:) A \ZZ_2(:,j,:)^T$;
    \State Compute the bottom $d_1$ eigenvectors of
$
A_2u_i=\lambda_iu_i;
$
    \State $U\gets [u_1,\dots,u_{d_1}]$;
  \Until{convergence or max
         number of iterations is met.}
\end{algorithmic}
\end{algorithm}  

Consider Algorithm~\ref{alg:alt1}.
Each update of $U$ reduces
the value of the objective function (\ref{eqn:ttraceA}),
so does the update of $V$.
The constraints $U^TU=I_{d_1}$ and $V^TV=I_{d_2}$
make the objective function (\ref{eqn:ttraceA}) bounded.
Hence Algorithm~\ref{alg:alt1} must converge.
However, it does not guarantee the convergence to the global minimum,
although each sub-problem can be solved optimally.
Indeed, the result depends on the initial $U$.
In practice, we simply set the initial $U$ as $I_{m_1}$.
Note that although $I_{m_1}$ is not of size $m_1$-by-$d_1$,
in the consequent computation the $U$ is of size $m_1$-by-$d_1$
and satisfies $U^TU=I_{d_1}$.

The methods in the second category,
i.e., 2D-PCA and GLRAM/CSA,
maximize (\ref{eqn:ttraceB}) subject to $U^TU=I_{d_1}$ and $V^TV=I_{d_2}$.
we can just substitute $-B$ for $A$ in Algorithm~\ref{alg:alt1}.
Equivalently, we can replace $A$ by $B$ and use the `top' eigenvectors
instead of the `bottom' eigenvectors.

Now we consider the last case,
minimizing (\ref{eqn:ttraceA}) and maximizing (\ref{eqn:ttraceB})
concurrently.
Following the discussion leading to (\ref{eqn:A1}),
we fix $U\in\IR^{m_1\times d_1}$. Then
maximizing (\ref{eqn:ttraceB}) in terms of $V\in\IR^{m_2\times d_2}$
is the same as maximizing $\trace(V^TB_1V)$, where
\begin{equation}
\label{eqn:B1}
B_1=\sum_{i=1}^{d_1}\ZZ_1(i,:,:) B\ZZ_1(i,:,:)^T,
\end{equation}
with $\ZZ_1=\XX\times_1 U^T$.
Now the sub-problem is to minimize
$\trace(V^TA_1V)$ and maximize $\trace(V^TB_1V)$ concurrently,
where $A_1$ and $B_1$ are defined in (\ref{eqn:A1}) and (\ref{eqn:B1}),
respectively.
If we enforce the orthogonality constraint $V^TV=I_{d_2}$,
then it is a trace ratio optimization problem
\cite{nbs:trace_ratio10,wyxth:trace_ratio07}.
For a better computational efficiency,
we consider the related but simpler program
\begin{equation}
\label{eqn:AB_prog1}
\left\{
    \begin{array}{ll} \displaystyle
    \mini_{V} & \displaystyle \trace(V^TA_1V) \\
    \textup{subject to}   & V^TB_1V=I_{d_2}.
    \end{array}
\right.
\end{equation}  
The minimizer $V\in\IR^{m_2\times d_2}$ of (\ref{eqn:AB_prog1})
consists of the bottom $d_2$ eigenvectors of the generalized eigenvalue
problem
$
A_1v_i=\lambda_iB_1v_i.
$

Likewise, if we fix $V$, then we will lead to the program
\begin{equation}
\label{eqn:AB_prog2}
\left\{
    \begin{array}{ll} \displaystyle
    \mini_{U} & \displaystyle \trace(U^TA_2U) \\
    \textup{subject to}   & U^TB_2U=I_{d_1},
    \end{array}
\right.
\end{equation}  
where
\begin{equation}
\label{eqn:B2}
B_2=\sum_{j=1}^{d_2} \ZZ_2(:,j,:) B \ZZ_2(:,j,:)^T
\end{equation}
with $\ZZ_2=\XX\times_2 V^T$.
We conclude an iterative algorithm
which solves (\ref{eqn:AB_prog1}) and (\ref{eqn:AB_prog2}) alternatively.
The pseudo-code is given in Algorithm~\ref{alg:alt2},
where we use the same initial $U$ in Algorithm~\ref{alg:alt1}.

\begin{algorithm}[H]
\caption{Alternating process \#2 for $U,V$.
\label{alg:alt2}}
\begin{algorithmic}
  \State {\bf input:} $\XX\in\IR^{m_1\times m_2\times n}$;
                      $A,B\in\IR^{n\times n}$.
  \State {\bf output:} projectors $U\in\IR^{m_1\times d_1}$, $V\in\IR^{m_2\times d_2}$.
  \State Set initial $U\gets I_{m_1}$;
  \Repeat
    \State $\ZZ_1 \gets \XX \times_1 U^T$;
    \State $\displaystyle A_1\gets \sum_{i=1}^{d_1} \ZZ_1(i,:,:)A\ZZ_1(i,:,:)^T$;
    \State $\displaystyle B_1\gets \sum_{i=1}^{d_1} \ZZ_1(i,:,:)B\ZZ_1(i,:,:)^T$;
    \State Compute the bottom $d_2$ eigenvectors of
$
A_1v_i=\lambda_iB_1v_i;
$
    \State $V\gets [v_1,\dots,v_{d_2}]$;
    \State $\ZZ_2 \gets \XX \times_2 V^T$;   
    \State $\displaystyle A_2\gets \sum_{j=1}^{d_2} \ZZ_2(:,j,:) A \ZZ_2(:,j,:)^T$;
    \State $\displaystyle B_2\gets \sum_{j=1}^{d_2} \ZZ_2(:,j,:) B \ZZ_2(:,j,:)^T$;
    \State Compute the bottom $d_1$ eigenvectors of
$
A_2u_i=\lambda_iB_2u_i;
$
    \State $U\gets [u_1,\dots,u_{d_1}]$;
  \Until{convergence or max \# iterations is met.}
\end{algorithmic}
\end{algorithm}

\subsection{Unilateral versus bilateral}

All the two-dimensional projections listed in Table~\ref{tbl:2dmethods_summary}
can be unilateral or bilateral.
The alternating processes given in Algorithms~\ref{alg:alt1} and \ref{alg:alt2}
are for bilateral projections.
The unilateral case is indeed simpler.
We just need to solve one eigenvalue or generalized eigenvalue problem, and
the optimal solution is reached with just one iteration.
In the literature,
\cite{czkl:2dlpp07,jwz:fisher2d06,ld:2dnpp07,nysp:laplacian2d08,yzfy:2dpca04}
use unilateral projections, whereas bilateral projections are adopted by
\cite{kwtlwv:g2dpca05,xyzlzu:csa08,ye:glram05,yjl:2dlda03}.

 The clear advantage of unilateral projections is
their computational efficiency.
On the other hand,  bilateral projections perform better in exploiting spatial
redundancy, and therefore need fewer dimensions for the projected space.
This is important in the application of data compression \cite{ye:glram05}.

Consider bilateral projections. 
GLRAM/CSA \cite{ye:glram05} and 2D-PCA \cite{kwtlwv:g2dpca05}
which maximize (\ref{eqn:ttraceB})
usually converge in 2 or 3 iterations in practice.
This is confirmed in our experiments.
Other methods which minimize (\ref{eqn:ttraceA})
require more iterations.
In the face recognition experiments, we limit the number of iterations to 5.
We found that more iterations usually help little in improving the recognition rate.

\subsection{Considerations for 2D-LDA \& 2D-LDA-R}
\label{sec:2dlda_comp}

There is a variant of Algorithm~\ref{alg:alt2}, described as follows.
Instead of (\ref{eqn:AB_prog1}),
another possibility is to maximize $\trace(V^TB_1V)$
subject to $V^TA_1V=I_{d_2}$.
This option leads to computing the top $d_2$ eigenvectors
of the generalized eigenvalue problem
\begin{equation}
\label{eqn:BAeig1}
B_1v_i=\lambda_iA_1v_i.
\end{equation}
Likewise, the other eigenvalue problem to solve
in the alternating process is
\begin{equation}
\label{eqn:BAeig2}
B_2u_i=\lambda_iA_2u_i.
\end{equation}

We use this variant to compute the 2D-LDA projections,
since we found that it usually achieves better performance
in face recognition.
However, it results in a problem in 2D-LDA-R
where the repulsion tensor is incorporated.
Note that solving the generalized eigenvalue problems
(\ref{eqn:BAeig1}) and (\ref{eqn:BAeig2}),
we need the matrices $A_1$ and $A_2$ being positive definite,
which is no longer guaranteed with repulsion.
The situation is even worse
in the case of bilateral projections due to the iteration.

In practice,
we set smaller penalty parameter $\beta$
for a modest repulsion power.
In addition, for bilateral projections,
we use only one iteration and in this iteration
$U$ and $V$ are computed independently,
i.e., like computing the unilateral projections
twice for $U$ and $V$, respectively.
These changes make 2D-LDA-R a useful method in face recognition.
However, instability occurs occasionally.
See Section~\ref{sec:fr_test}
for an example in the experiment on the {\tt UMIST} face images.

\subsection{Pre-processing and post-processing}
\label{sec:prepost}

In LDA, LPP, NPP and their variants,
it is common to pre-process the training data matrix by PCA,
in order to avoid a singular matrix in the eigenvalue computation.
Likewise, in 2D-LDA, 2D-LPP, 2D-NPP and their variants,
one can pre-process the data by 2D-PCA.
This can sometimes improve the stability,
explained as follows.

Consider Algorithms~\ref{alg:alt1} and \ref{alg:alt2}.
We would like to ensure that
the matrices $A_1,B_1\in\IR^{m_2\times m_2}$ and
$A_2,B_2\in\IR^{m_1\times m_1}$ are of full rank.
We concentrate on $A_1$.
The discussion for $B_1,A_2,B_2$ is similar.
The rank of $A_1$
is at most $d_1\min\{m_2,\rank(A)\}$.
Hence if $d_1\rank(A)<m_2$, we are guaranteed
a singular $A_1$.
It is known that in supervised mode,
$\rank(A)\leq n-c$,
where $n$ is the number of training images and
$c$ is the number of classes
\cite{hyunz:laplacian05,ks:onpp05,ks:onpp07}.
In practice, without the singularity problem,
good performance can be achieved with very small $d_1$ or $d_2$.
Hence when the image size is large or
the number of training images is small,
pre-processing by 2D-PCA helps improve the stability.
We found that it is an issue in the experiments
on the {\tt AR} database.

Another approach is to post-process
 the projected data by a one-dimensional method,
in order to improve data compression or recognition performance.
 Examples include GLRAM + SVD \cite{ye:glram05},
2D-LPP + PCA \cite{czkl:2dlpp07},
and 2D-LDA + LDA \cite{yjl:2dlda03}.
For the sake of simplicity, 
we did not adopt this approach in the experiments.
The two directions open a door to numerous composite projections
which deserve further investigations.

\section{Experimental results}
\label{sec:exp}

The
experiments were performed in MATLAB on a PC equipped
with a four-core Intel Xeon E5504 @ 2.0GHz processor
and 4GB memory. 
We use 4 face image databases,
{\tt ORL} faces \cite{sh:orl94},
{\tt UMIST} faces \cite{ga:umist98},
{\tt AR} faces \cite{mk:pca_lda01}, and
{\tt ESSEX} faces \cite{spacek:essex02}.
We compare empirically the performance of the repulsion methodology
(see Section~\ref{sec:trepulsion}),
the existing two-dimensional methods
(see Section~\ref{sec:2dmethods}),
and the classical linear dimensionality methods
using image-as-vector representation
(see Appendix~\ref{sec:1dmethods}).

We report the results of 2D-PCA, 2D-LDA, 2D-PCA-R and 2D-LDA-R,
as well as the results of 2D-LPP, 2D-NPP, 2D-OLPP-R, and 2D-ONPP-R.
In practice, we found that
2D-LPP and 2D-NPP are generally better options than
2D-OLPP and 2D-ONPP.
On the other hand,
2D-OLPP-R and 2D-ONPP-R often outperform
2D-LPP-R and 2D-NPP-R.
Hence we omit the results of
2D-OLPP, 2D-ONPP, 2D-LPP-R, and 2D-NPP-R
in this section.

Each recognition rate reported is the average
from 20 random realizations of the training images, and
the rest images are used for testing.
In Tables~\ref{tbl:orl_best}--\ref{tbl:essex_best},
we list the best recognition rate and the corresponding dimension
for each method and each database.
The statistics of the one-dimensional methods
are from \cite{ks:repulsion09},
except for the NPP and LDA-R methods, whose numbers are obtained
from the recent experiments.

\subsection{Face image databases}
\label{sec:face_database}

\begin{figure}[htb]
\begin{center}
\resizebox{0.6\textwidth}{!}{
\includegraphics[width=\textwidth]{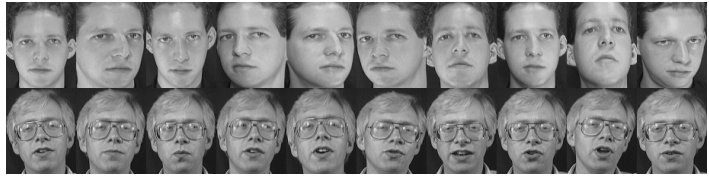}
}
\caption{Sample {\tt ORL} face images.
\label{fig:orl_faces}}
\end{center}
\end{figure}

The {\tt ORL} (Olivetti Research Laboratory) face database \cite{sh:orl94}
contains images of 40 subjects.
Each subject has 10 grayscale images of size 112-by-92
with various facial expressions (smiling/non-smiling, etc.).
In total there are 400 images.
Sample face images of the first two individuals
are displayed in Figure~\ref{fig:orl_faces}.

\begin{figure}[htb]
\begin{center}
\resizebox{0.6\textwidth}{!}{
\includegraphics[width=\textwidth]{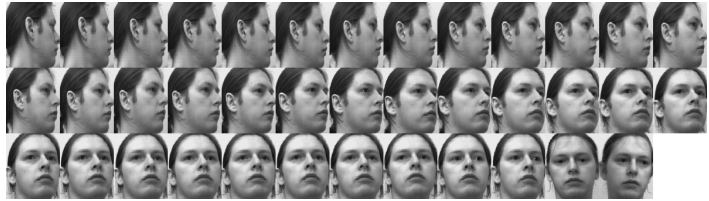}
}
\caption{Sample {\tt UMIST} face images.
\label{fig:umist_faces}}
\end{center}
\end{figure}

The {\tt UMIST} database contains grayscale images of 20 subjects.
We use a set of cropped images.
Each subject has from 19 to 48 images of size 112-by-92.
In total there are 575 images.
Figure~\ref{fig:umist_faces} gives sample images of
the first individual.

\begin{figure}[htb]
\begin{center}
\resizebox{0.6\textwidth}{!}{
\includegraphics[width=\textwidth]{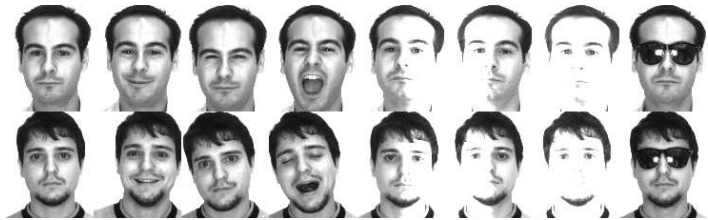}
}
\caption{Sample {\tt AR} face images.
\label{fig:ar_faces}}
\end{center}
\end{figure}

We use the {\tt AR} database \cite{mk:pca_lda01}
which contains 1,008 grayscale 768-by-576 images of 126 subjects,
8 images per subject.
These images were cropped and downsized to be 224-by-184
before the recognition is performed in the experiments.
Figure~\ref{fig:ar_faces} presents the sample images of the first two
subjects.

We also use the {\tt ESSEX} database \cite{spacek:essex02}.
The 95' version contains 1,440 color 200-by-180 images of 72 subjects.
Each subject has 20 images.
We converted these color images to grayscale
so they can be represented as matrices.

\subsection{Performance sensitivity to repulsion}
\label{sec:kbeta}

There are two parameters to select for the repulsion technique.
One is the number $k$  of neighbors per vertex,
used for constructing a repulsion graph.
The other is the penalty parameter $\beta$,
which determines the strength of the repulsion term.

\begin{figure*}[htbp]
\begin{center}
\resizebox{0.9\textwidth}{!}{ 
\includegraphics{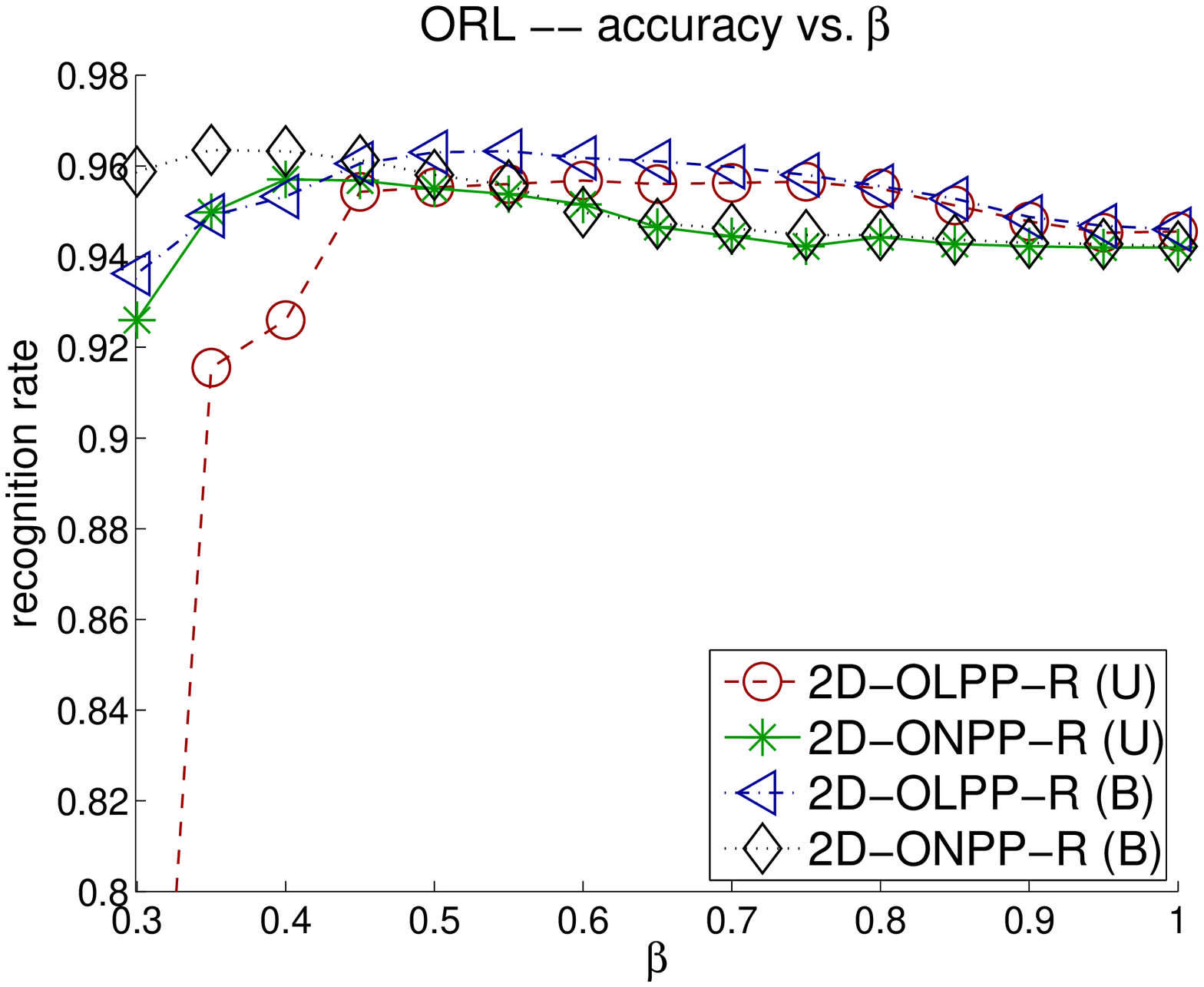}
\hspace{1in}
\includegraphics{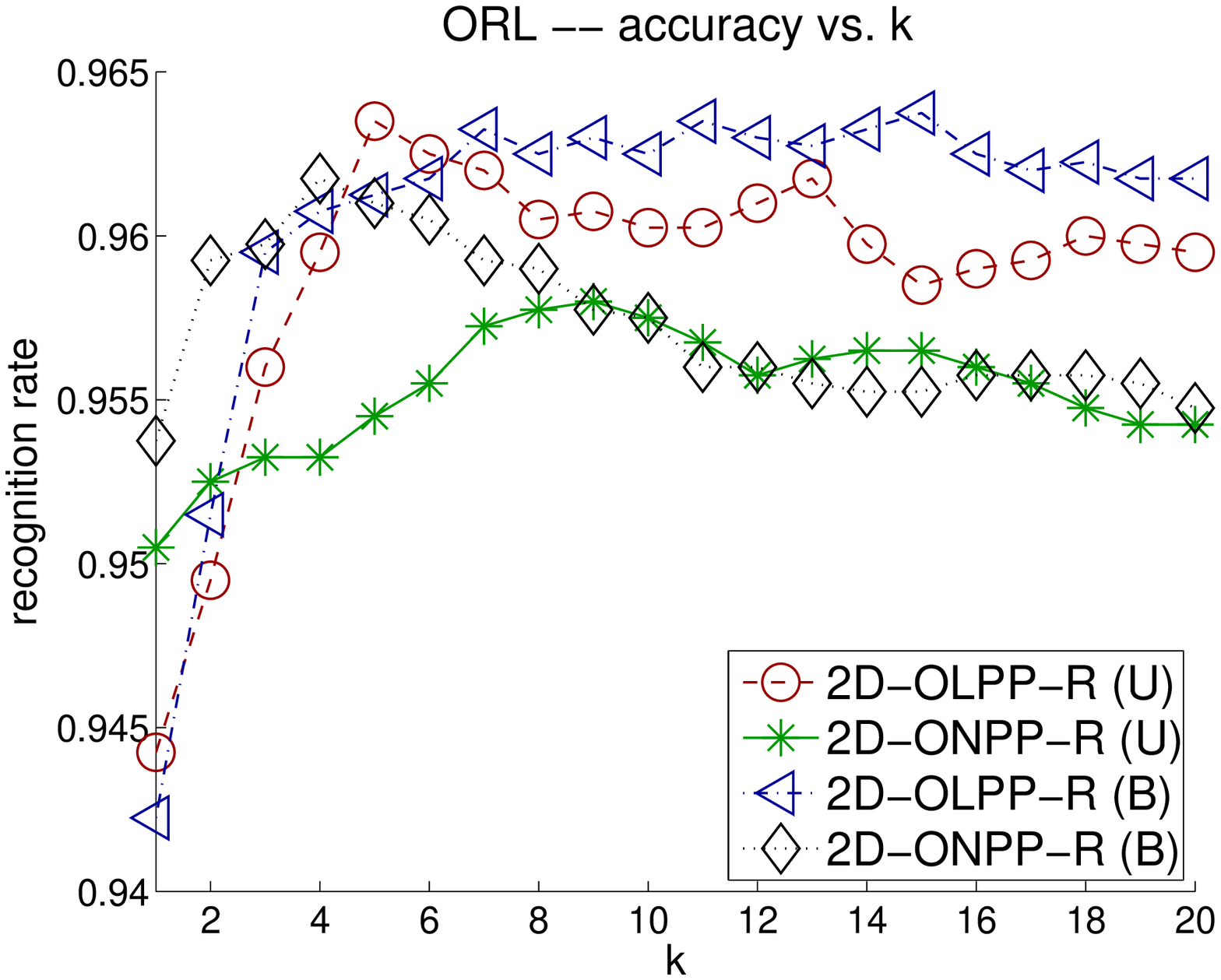}
}
\caption{Sensitivity of performance to varying $\beta$ (left) and $k$ (right)
for the {\tt ORL} database.
\label{fig:kbeta}}
\end{center}
\end{figure*}

To examine the sensitivity of the recognition performance
to these parameters, we conducted two experiments as follows.
We use the 2D-ONPP-R and 2D-OLPP-R methods, in both
unilateral and bilateral projections.
For unilateral projection,
we set $d_2=10$ and the result is marked by `(U)'.
For bilateral projection,
we set $d_1=d_2=10$ and the result is marked by `(B)'.
We use the images in the {\tt ORL} database, and
set the number of training images per class as 5.
In one experiment, we set $k=6$ and $\beta=0.3,0.35,\dots,1$.
In the other experiment, we set $k=1,2,\dots,20$ and $\beta=0.5$.
The results are shown in Figure~\ref{fig:kbeta}.

It can be observed from Figure~\ref{fig:kbeta} that
the recognition rate changes modestly
as long as $\beta$ are $k$ are large enough.

\subsection{Face recognition results}
\label{sec:fr_test}

We report the results for 4 databases:
{\tt ORL}, {\tt UMIST}, {\tt AR}, and {\tt ESSEX}.
Regarding the repulsion parameters,
we use $k=6$ in all cases.
In both 2D-OLPP-R and 2D-ONPP-R,
we set $\beta=0.5$, but for the {\tt ESSEX} database,
we set $\beta=1.0$.
As discussed in Section~\ref{sec:2dlda_comp},
we need a smaller $\beta$ for 2D-LDA-R,
in which we set $\beta=0.2$.

\begin{figure*}[htbp]
\begin{center}
\resizebox{0.9\textwidth}{!}{ 
\includegraphics{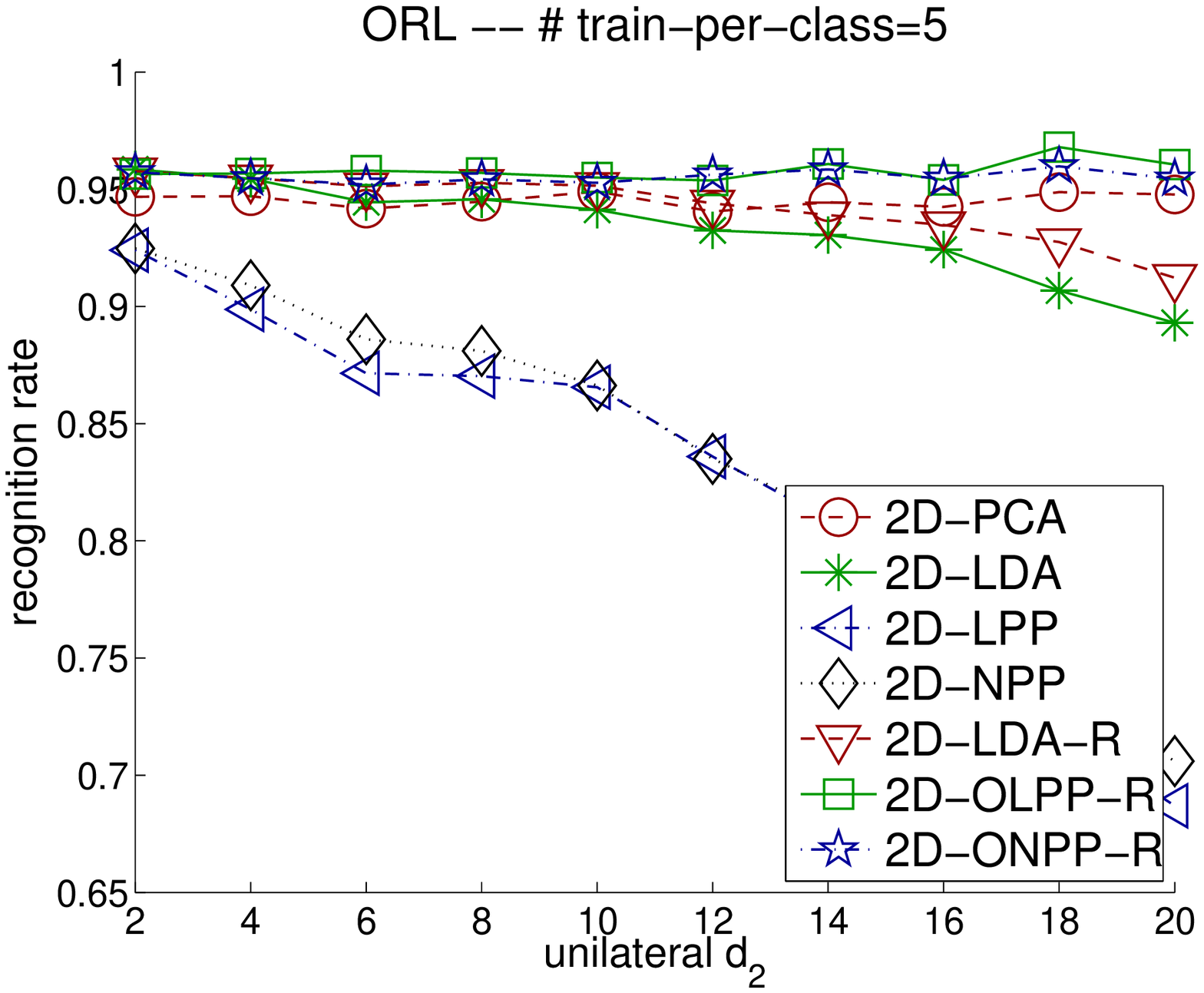}
\hspace{1in}
\includegraphics{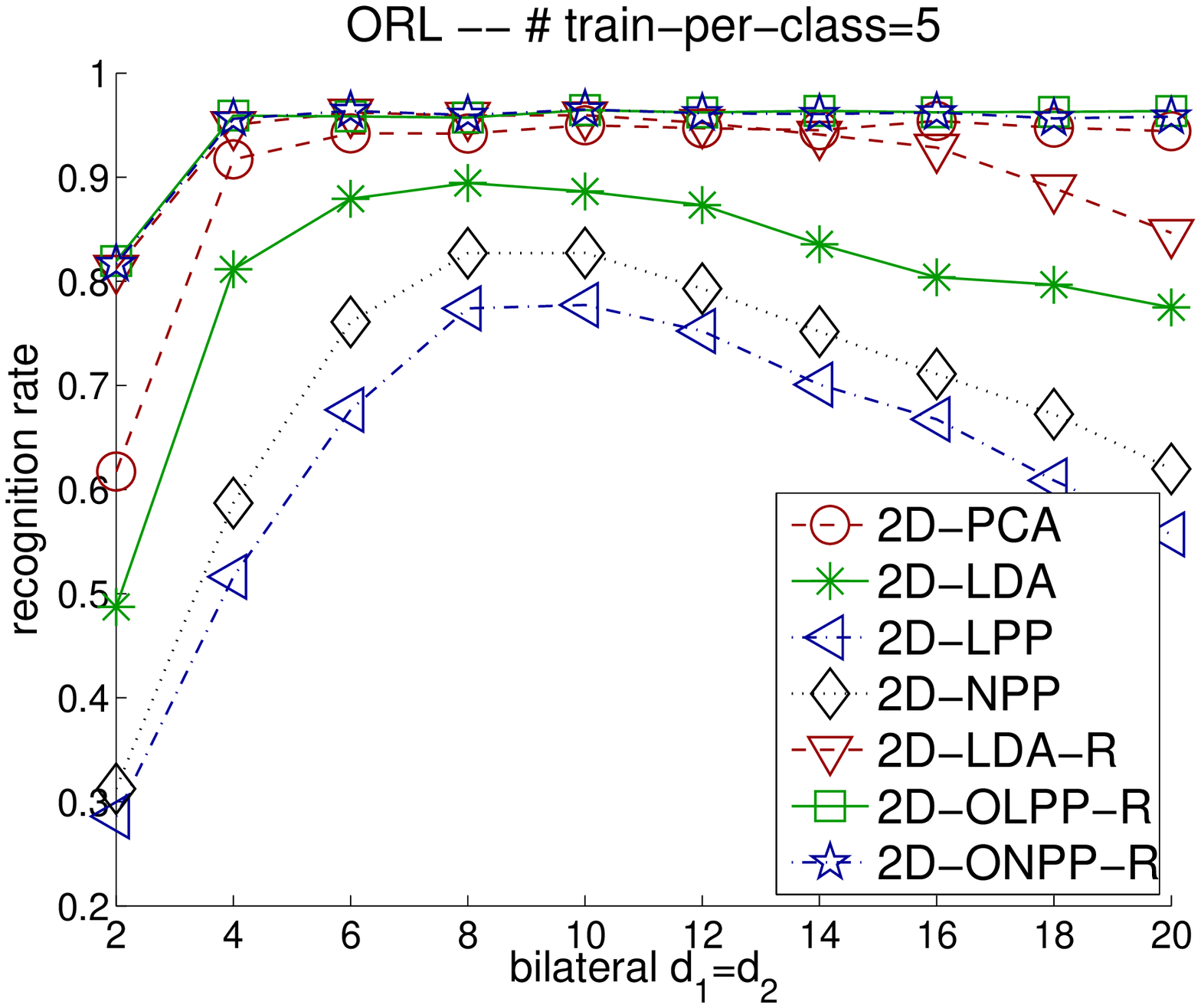}
}
\caption{Accuracy versus dimension for the {\tt ORL} database;
unilateral projection (left) and bilateral projection (right).
\label{fig:orl_plot}}
\end{center}
\end{figure*}

The results for the {\tt ORL} database are displayed in Figure~\ref{fig:orl_plot},
where the number of training images per class is set as 5.
The left plot shows the recognition rates with the unilateral projections
with the column dimension $d_2=2,4,\dots,20$,
whereas the right plot shows the recognition rates with
the bilateral projections
with the row/column dimension $d_1=d_2=2,4,\dots,20$.

Two observations from Figure~\ref{fig:orl_plot} can be made.
\begin{enumerate}
\item
In both settings of unilateral and bilateral projections,
2D-LDA-R improves 2D-LDA, and
2D-OLPP-R and 2D-ONPP-R outperform 2D-LPP and 2D-NPP handsomely.
\item
2D-OLPP-R is the best option for the {\tt ORL} database,
whereas 2D-ONPP-R is the close runner-up.
\end{enumerate}

Table~\ref{tbl:orl_best} lists the best recognition rates and
the corresponding dimensions,
using the 7 two-dimensional (2D) methods and their one-dimen-sional (1D) peers.
On average, the 1D methods and 2D methods are
comparable in performance.
However, the 1D methods with repulsion Laplaceans
perform slightly better than the 2D methods with repulsion
tensors.

\begin{table*}[htbp]
\center{
\caption{Best achieved error rates and the corresponding dimensions
for the {\tt ORL} database.\label{tbl:orl_best}}
\begin{tabular}{l|cc||l|cc|cc} \hline
\multicolumn{3}{c||}{1D Method} & \multicolumn{5}{c}{2D Method} \\ \hline
\multirow{2}{*}{method} & \multirow{2}{*}{\# dim.} & \multirow{2}{*}{error} &
\multirow{2}{*}{method} & \multicolumn{2}{c|}{unilateral} & \multicolumn{2}{c}{bilateral} \\
 &  &  &  & \# dim. & error & \# dim. & error  \\ \hline
PCA    & 90 & 5.12\% & 2D-PCA    & 10 & 5.10\% & 16 & 4.60\% \\
LDA    & 65 & 6.95\% & 2D-LDA    & 2  & 4.15\% & 8  & 10.6\% \\
LPP    & 60 & 10.8\% & 2D-LPP    & 2  & 7.60\% & 10 & 22.3\% \\
NPP    & 15 & 9.50\% & 2D-NPP    & 2  & 7.53\% & 10 & 17.3\% \\
LDA-R  & 70 & 2.90\% & 2D-LDA-R  & 2  & 4.23\% & 6  & 3.78\% \\
OLPP-R & 55 & 2.82\% & 2D-OLPP-R & 18 & 3.20\% & 10 & 3.55\% \\
ONPP-R & 90 & 3.40\% & 2D-ONPP-R & 18 & 4.03\% & 10 & 3.50\% \\
\hline
\end{tabular}
}
\end{table*}

\begin{figure*}[htbp]
\begin{center}
\resizebox{0.9\textwidth}{!}{ 
\includegraphics{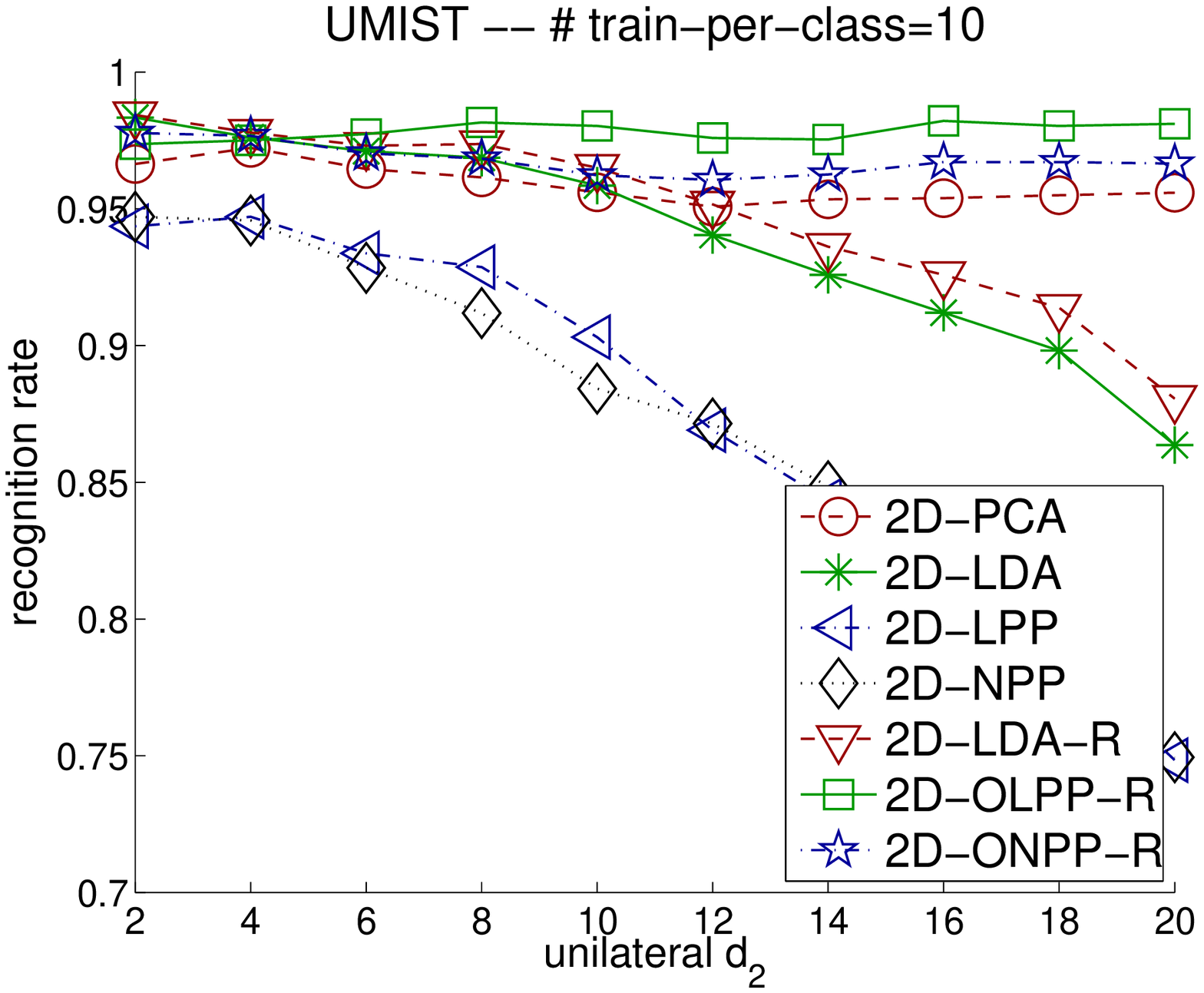}
\hspace{1in}
\includegraphics{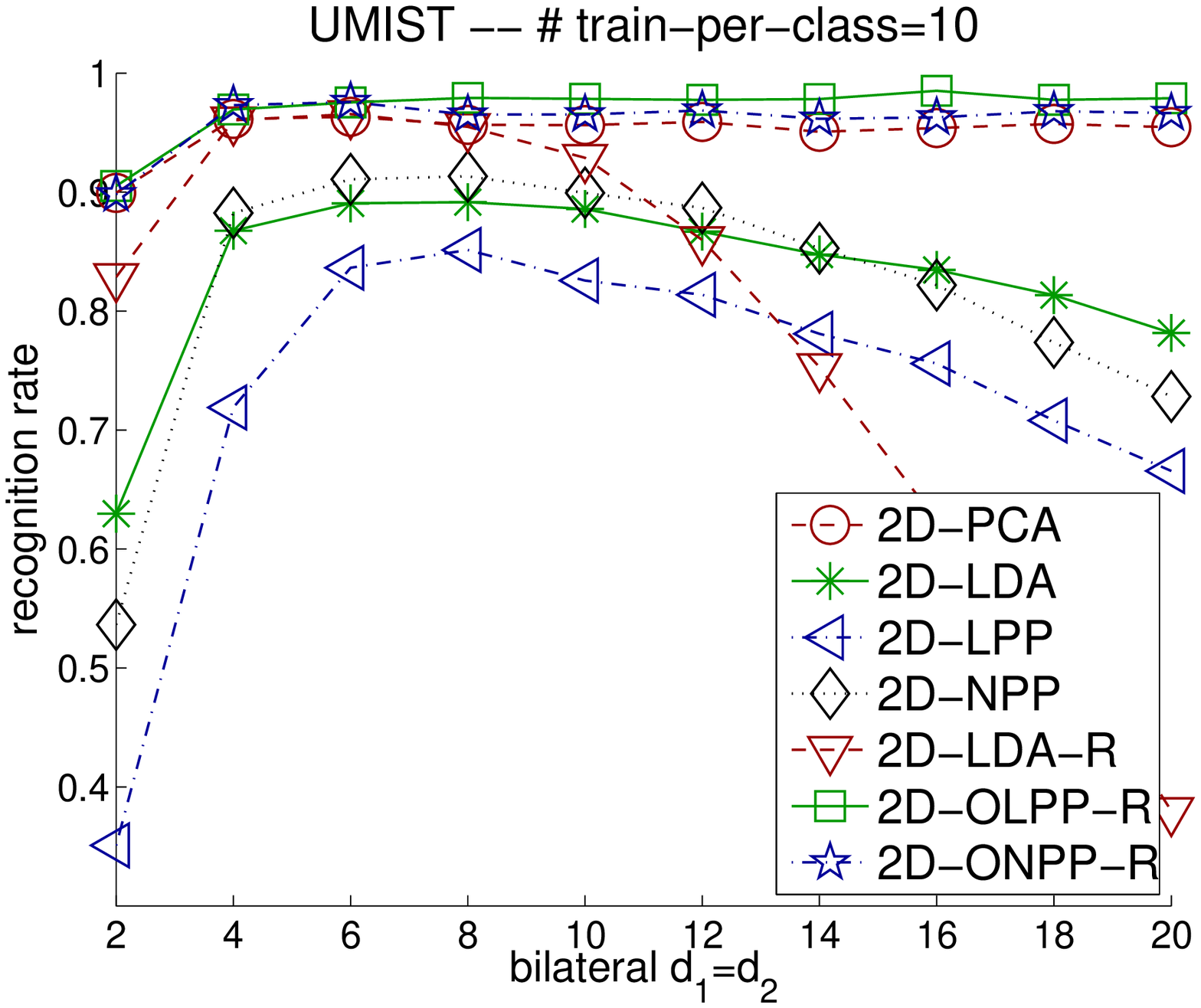}
}
\caption{Accuracy versus dimension for the {\tt UMIST} database;
unilateral projection (left) and bilateral projection (right).
\label{fig:umist_plot}}
\end{center}
\end{figure*}

\vspace{0.5cm}
Figure~\ref{fig:umist_plot} displays the results for the {\tt UMIST}
database, where the number of training images is set as 10.
Table~\ref{tbl:umist_best} lists the best recognition rates and   
the corresponding dimensions.
The observations made previously for the {\tt ORL} image set
are also valid here, except for that the performance of
the 2D-LDA-R bilateral projection deteriorates quickly
while the dimension increases.
An explanation of this instability is given
in Section~\ref{sec:2dlda_comp}.
Among the two-dimensional methods,
both 2D-OLPP-R and 2D-ONPP-R are the apparent winners
for the {\tt UMIST} and {\tt ORL} databases.

\begin{table*}[htbp]
\center{
\caption{Best achieved error rates and the corresponding dimensions
for the {\tt UMIST} database.\label{tbl:umist_best}}
\begin{tabular}{l|cc||l|cc|ccccc} \hline
\multicolumn{3}{c||}{1D Method} & \multicolumn{5}{c}{2D Method} \\ \hline
\multirow{2}{*}{method} & \multirow{2}{*}{\# dim.} & \multirow{2}{*}{error} &
\multirow{2}{*}{method} & \multicolumn{2}{c|}{unilateral} & \multicolumn{2}{c}{bilateral} \\
 &  &  &  & \# dim. & error & \# dim. & error  \\ \hline
PCA    & 65 & 4.12\% &  2D-PCA    & 4  & 2.77\% & 6  & 3.67\% \\
LDA    & 30 & 3.44\% &  2D-LDA    & 2  & 1.67\% & 8  & 1.07\% \\
LPP    & 10 & 4.03\% &  2D-LPP    & 4  & 5.29\% & 8  & 1.48\% \\
NPP    & 15 & 4.56\% &  2D-NPP    & 2  & 5.29\% & 8  & 8.66\% \\
LDA-R  & 95 & 1.62\% &  2D-LDA-R  & 2  & 1.57\% & 6  & 3.44\% \\
OLPP-R & 30 & 0.93\% &  2D-OLPP-R & 16 & 1.79\% & 16 & 1.48\% \\
ONPP-R & 15 & 1.45\% &  2D-ONPP-R & 2  & 2.23\% & 6  & 2.42\% \\
\hline
\end{tabular}
}
\end{table*}


\begin{figure*}[htbp]
\begin{center}
\resizebox{0.9\textwidth}{!}{ 
\includegraphics{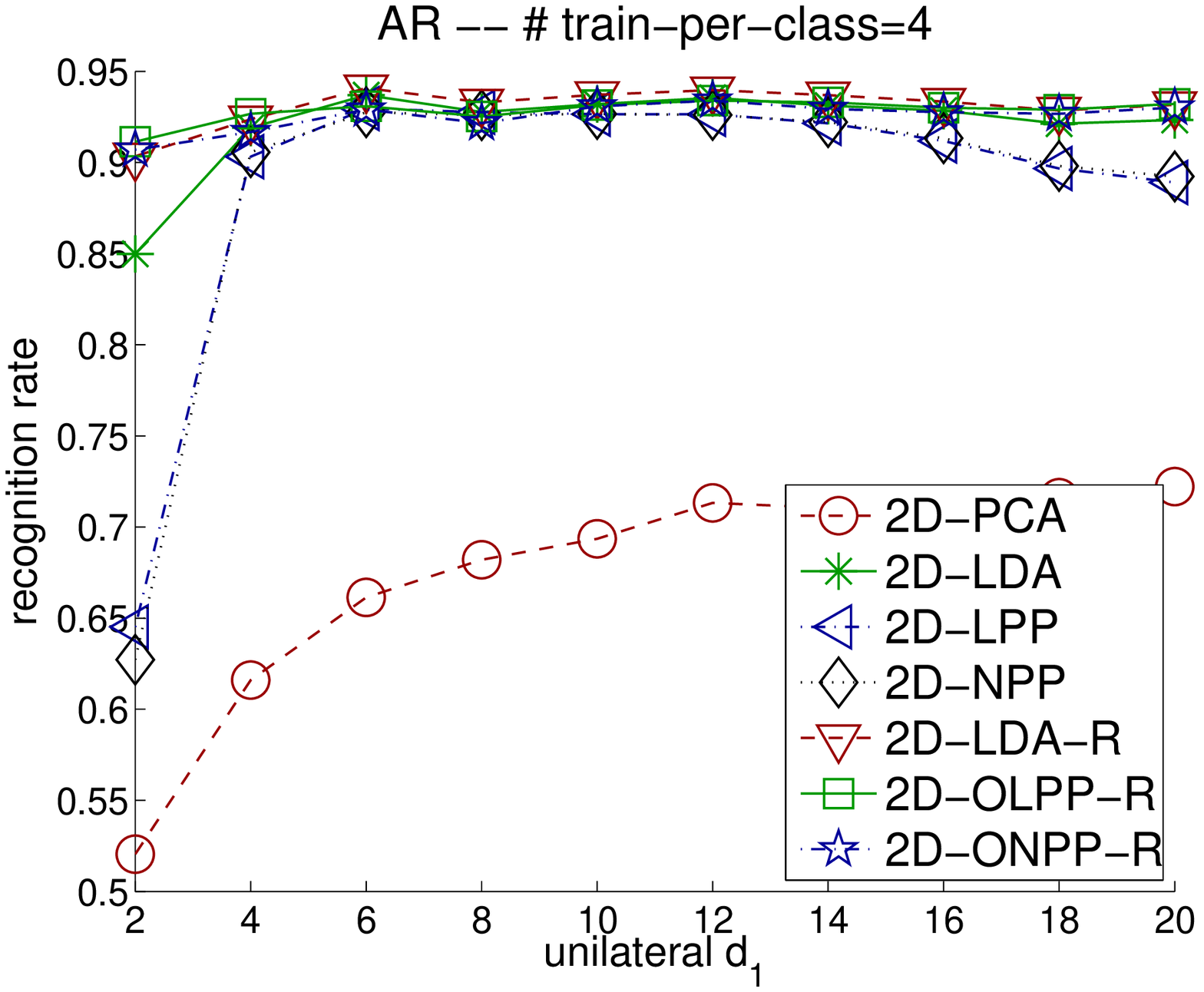}
\hspace{1in}
\includegraphics{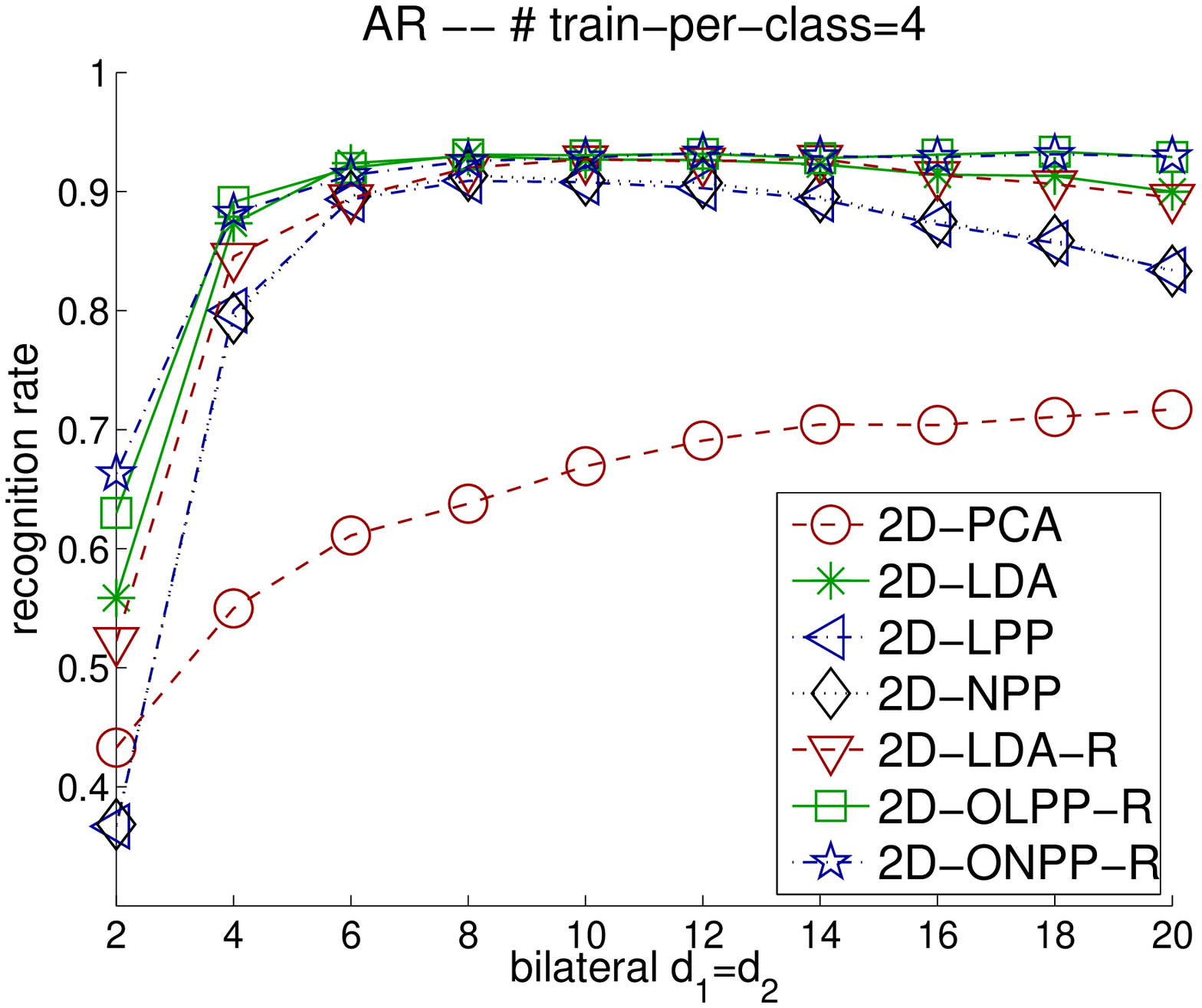}
}
\caption{Accuracy versus dimension for the {\tt AR} database;
unilateral projection (left) and bilateral projection (right).
\label{fig:ar_plot}}
\end{center}
\end{figure*}

\begin{table*}[htbp]
\center{
\caption{Best achieved error rates and the corresponding dimensions
for the {\tt AR} database.\label{tbl:ar_best}}
\begin{tabular}{l|cc||l|cc|ccccc} \hline
\multicolumn{3}{c||}{1D Method} & \multicolumn{5}{c}{2D Method} \\ \hline
\multirow{2}{*}{method} & \multirow{2}{*}{\# dim.} & \multirow{2}{*}{error} &
\multirow{2}{*}{method} & \multicolumn{2}{c|}{unilateral} & \multicolumn{2}{c}{bilateral} \\
 &  &  &  & \# dim. & error & \# dim. & error  \\ \hline
PCA    & 100 & 28.6\% &  2D-PCA    & 20 & 27.8\% & 20 & 28.3\% \\
LDA    & 100 & 10.3\% &  2D-LDA    & 6  & 6.31\% & 8  & 7.03\% \\
LPP    & 15  & 7.12\% &  2D-LPP    & 6  & 7.04\% & 8  & 9.10\% \\
NPP    & 20  & 7.60\% &  2D-NPP    & 6  & 7.24\% & 8  & 8.66\% \\
LDA-R  & 85  & 5.98\% &  2D-LDA-R  & 6  & 5.92\% & 14 & 7.25\% \\
OLPP-R & 75  & 3.96\% &  2D-OLPP-R & 12 & 6.61\% & 18 & 6.66\% \\
ONPP-R & 45  & 4.04\% &  2D-ONPP-R & 12 & 6.61\% & 12 & 6.76\% \\
\hline
\end{tabular}
}
\end{table*}

The results for the {\tt AR} database are illustrated in
Figure~\ref{fig:ar_plot},
where we set the number of training images as 4.
Note that except for 2D-PCA,
we pre-process the data by 2D-PCA, explained in Section~\ref{sec:prepost}.
Table~\ref{tbl:ar_best} lists the best recognition rates and   
the corresponding dimensions.
2D-ONPP-R and 2D-OLPP-R are still the best two-dimensional methods in performance.
In addition, 2D-LDA-R with unilateral projection performs as good in this case.
It is interesting to note that 2D-LPP and 2D-NPP yield very similar results.
Also note that PCA and 2D-PCA does not perform that well for this case.

\begin{figure*}[htbp]
\begin{center}
\resizebox{0.9\textwidth}{!}{ 
\includegraphics{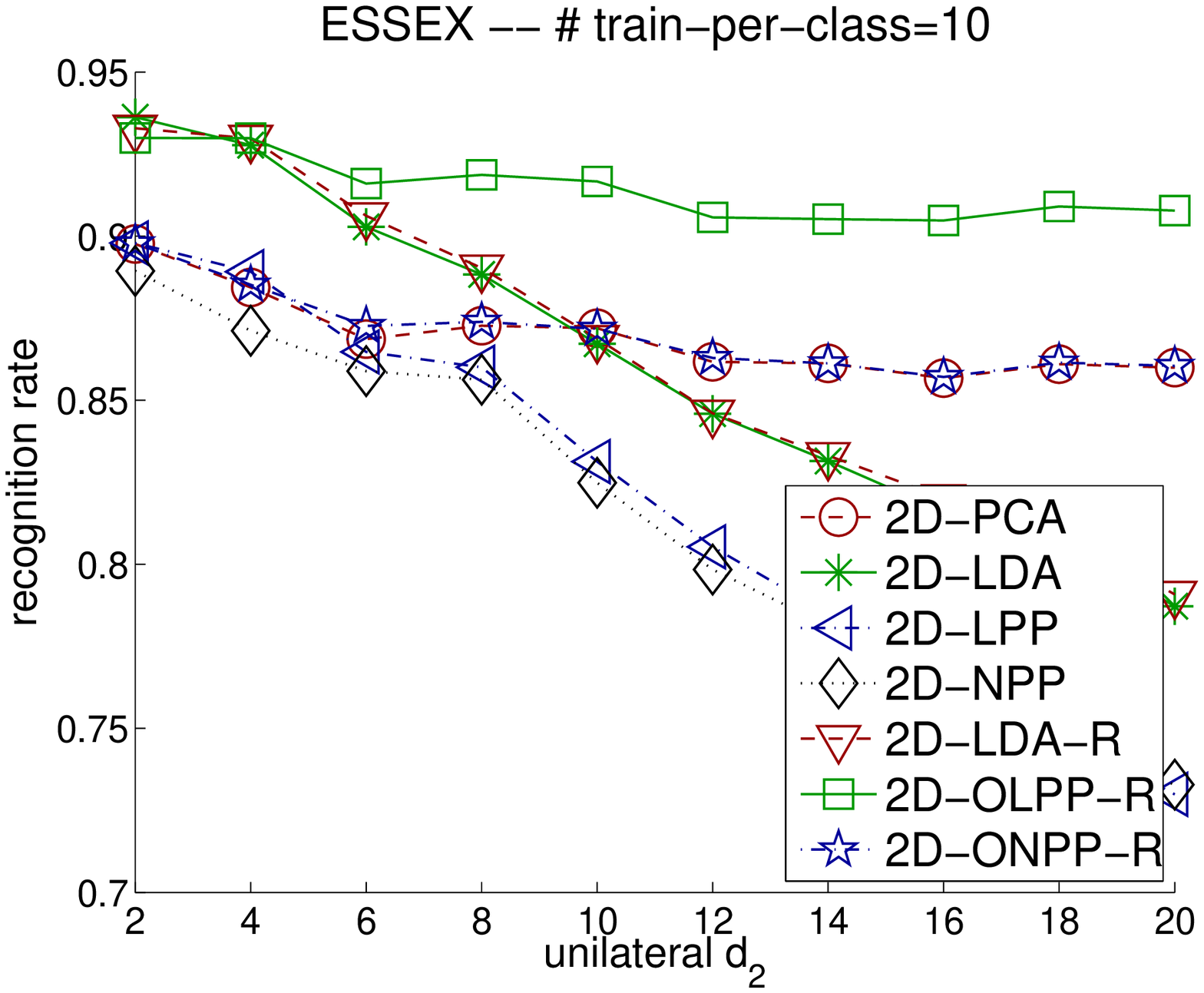}
\hspace{1in}
\includegraphics{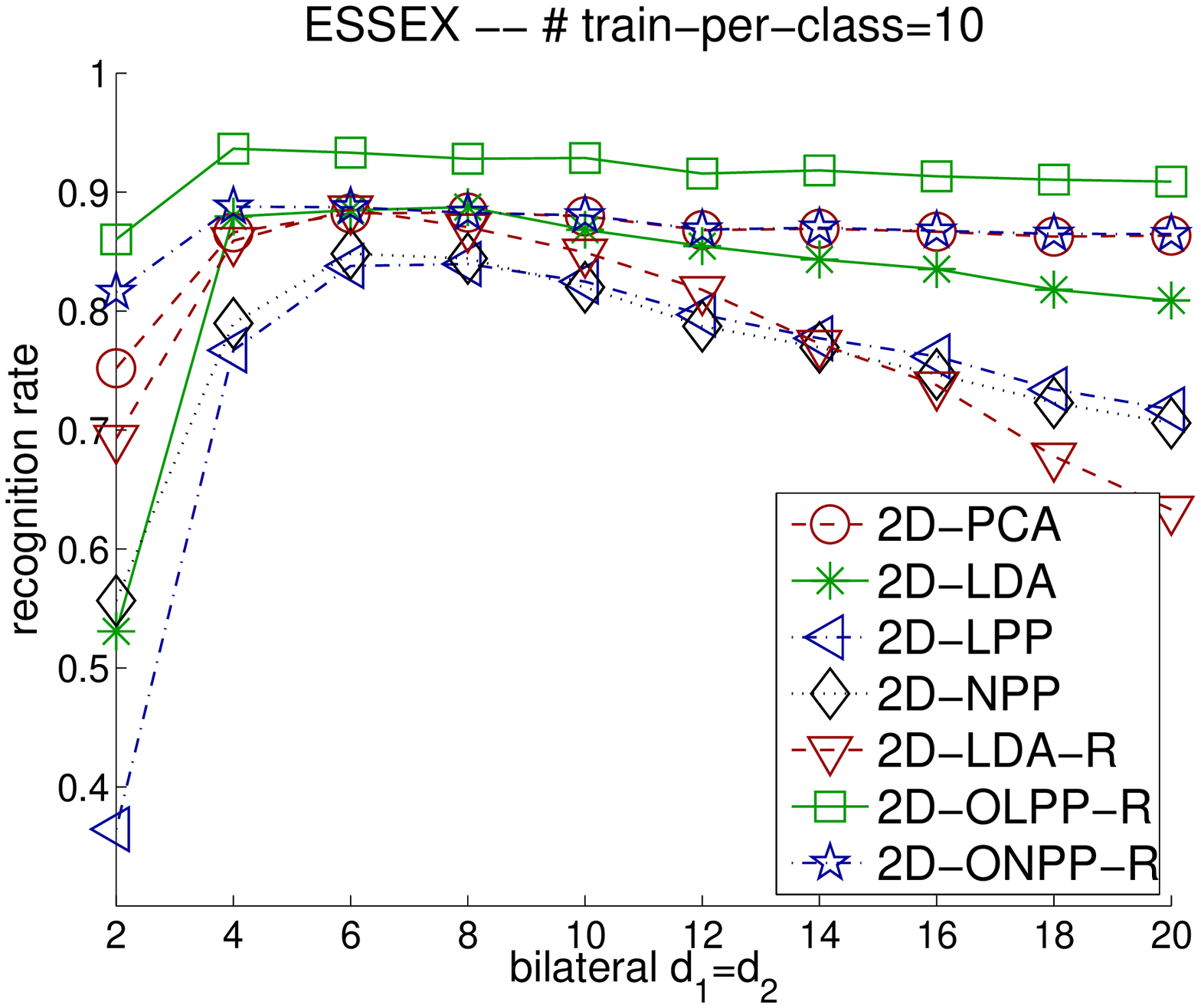}
}
\caption{Accuracy versus dimension for the {\tt ESSEX} database;
unilateral projection (left) and bilateral projection (right).
\label{fig:essex_plot}}
\end{center}
\end{figure*}

Figure~\ref{fig:essex_plot} displays the results for the {\tt ESSEX} database,
where we set the number of training images as 10.
Table~\ref{tbl:essex_best} lists the best recognition rates and
the corresponding dimensions.
In this case, 2D-OLPP-R outperforms the other methods,
including all 1D methods.
This database is the hardest one tested in \cite{ks:repulsion09}.
It hints the potential of the repulsion tensors
for challenging classification tasks.

\begin{table*}[t]
\center{
\caption{Best achieved error rates and the corresponding dimensions
for the {\tt ESSEX} database.\label{tbl:essex_best}}
\begin{tabular}{l|cc||l|cc|ccccc} \hline
\multicolumn{3}{c||}{1D Method} & \multicolumn{5}{c}{2D Method} \\ \hline
\multirow{2}{*}{method} & \multirow{2}{*}{\# dim.} & \multirow{2}{*}{error} &
\multirow{2}{*}{method} & \multicolumn{2}{c|}{unilateral} & \multicolumn{2}{c}{bilateral} \\
 &  &  &  & \# dim. & error & \# dim. & error  \\ \hline
PCA    & 65 & 15.1\% &  2D-PCA    & 2  & 10.2\% & 8  & 11.7\% \\
LDA    & 55 & 61.6\% &  2D-LDA    & 2  & 6.38\% & 8  & 11.3\% \\
LPP    & 10 & 15.3\% &  2D-LPP    & 2  & 10.2\% & 8  & 16.0\% \\
NPP    & 15 & 17.3\% &  2D-NPP    & 2  & 11.1\% & 6  & 15.2\% \\
LDA-R  & 20 & 10.4\% &  2D-LDA-R  & 2  & 6.72\% & 6  & 11.4\% \\
OLPP-R & 25 & 12.5\% &  2D-OLPP-R & 2  & 7.01\% & 4  & 6.35\% \\
ONPP-R & 65 & 9.88\% &  2D-ONPP-R & 2  & 10.2\% & 4  & 11.2\% \\
\hline
\end{tabular}
}
\end{table*}

In addition to the attractive recognition rates
by 2D-OLPP-R and 2D-ONPP-R.
It is worth noting that the performance of 2D-OLPP-R and 2D-ONPP-R
is very insensitive to the variation of the reduced dimension
in all our tests.
This is an appealing feature, especially considering that
in real applications we may not have the labels of test images.

\subsection{1D methods versus 2D methods}
\label{sec:1d_vs_2d}

One-dimensional (1D) methods use the image-as-vector representation,
whereas two-dimensional (2D) methods use the image-as-matrix representation.
This fundamental difference governs the computation.
Whether 2D methods or 1D methods are faster
depends on several factors.
For example, size of the images, number of training faces,
how the algorithms are implemented, and the numerical libraries used, etc.

The high level idea is that
1D methods, which vectorize images,
need to solve a much larger eigenvalue problem.
On the other hand,
the computation of the 2D methods
is often dominated by forming the eigenvalue problem,
i.e., computing (\ref{eqn:A1}), (\ref{eqn:A2}), (\ref{eqn:B1}),
and (\ref{eqn:B2}).
In practice, this part is usually less expensive than the
eigenvalue computation of the 1D method.
Indeed, it has been reported that 2D methods
are more efficient than their 1D counterparts,
e.g., \cite{czkl:2dlpp07,nysp:laplacian2d08,yzfy:2dpca04,ye:glram05,yjl:2dlda03}.
We have also observed this property in our experiments.
Note however that one can downsize the image matrices
to reduce the cost of the eigenvalue computation.

\section{Conclusions and future work}
\label{sec:end}

We have developed a methodology with repulsion tensors
to enhance the two-dimensional projection methods for classification.
It can be regarded as a multilinear generalization of
a technique called repulsion Laplaceans \cite{ks:repulsion09}.
The key idea is improve the projection
by repelling data items
which are from different classes 
but close to each other in the input space.
The experiments on face recognition exhibit
significant improvement of recognition performance
by the proposed technique.

Some two-dimensional projection methods can be formulated as
a tensor trace ratio optimization problem.
Instead of solving this problem directly,
we use Algorithm~\ref{alg:alt2} as an efficient workaround.
It may be worth developing efficient algorithms
for the tensor trace ratio optimization problem.

All the methods presented in this paper are linear.
Hence it may fail to discover the underlying latent structure
if the image manifold is highly nonlinear.
A research topic is
how to learn the nonlinear structure
of images or higher order tensor objects
in the supervised setting.

\section*{Acknowledgement}

Supported by grant NSF/CCF-1318597,
this work was mainly done in 2012,
while the author was a research associate in the University of Minnesota.
Thanks to Yousef Saad for his support and
his help in editing the manuscript.


\bibliographystyle{plain}
\bibliography{face}

\appendix

\section{Basics of tensors}
\label{sec:tensor_intro}

An $r$th order {\em tensor} is an array of $r$ indices.
A vector is a first order tensor, and
a matrix is a second order tensor.
Different dimensions are also called {\em modes}.
To facilitate the presentation,
we will illustrate with third and fourth order tensors
with specific modes.
However, the definitions for higher order tensors
in general modes
can be straightforwardly extended.
The readers are referred to \cite{bk:tensor06} for details.

Let $\AA,\BB\in\IR^{I\times J\times K}$ be third order tensors,
where $I,J,K$ are positive integers.
The inner product $\langle\AA,\BB\rangle$ of tensors $\AA,\BB$ is defined by
\begin{equation}
\label{eqn:tensor_inner_product}
\langle\AA,\BB\rangle
=
\sum_{i=1}^I\sum_{j=1}^J\sum_{k=1}^K a_{ijk}b_{ijk}.
\end{equation}
Two tensors $\AA,\BB$ are {\em orthogonal} to each other
if $\langle\AA,\BB\rangle=0$.
The Frobenius norm of a tenor $\AA$ is defined by
\begin{equation}
\label{eqn:tensor_norm}
\|\AA\|=\sqrt{\langle\AA,\AA\rangle}.
\end{equation}

We can transform a higher order tensor to a matrix
by merging dimensions and rearranging the elements.
This is called `unfolding' \cite{lmv:hosvd00},
`flattening' \cite{vt:tensorfaces02},
or `matricizing' \cite{bk:tensor06} a tensor.

For example, 
given a third order tensor $\AA=[a_{ijk}]\in\IR^{I\times J\times K}$,
we can have
\begin{eqnarray*}
A_{(1)}=[a_{ip}^{(1)}]\in\IR^{I\times JK},
 &  a_{ijk}=a_{ip}^{(1)},  &  p=j+(k-1)J,\\
A_{(2)}=[a_{jp}^{(2)}]\in\IR^{J\times IK},
 &  a_{ijk}=a_{jp}^{(2)},  &  p=k+(i-1)K,\\
A_{(3)}=[a_{kp}^{(3)}]\in\IR^{K\times IJ},
 &  a_{ijk}=a_{kp}^{(3)},  &  p=i+(j-1)I,
\end{eqnarray*}
where by default we have used the forward cyclic ordering.
The other option of ordering is
backward cyclic \cite{bk:tensor06}.
We use $A_{(1;2,3)}$ and $A_{(1;3,2)}$ to specify
the forward and backward cyclic orderings, respectively.
Here is another example to matricize a fourth order tensor
$\AA=[a_{ijkh}]\in\IR^{I\times J\times K\times H}$ as
\begin{equation}
\label{eqn:matricize4}
A_{(1,2;3,4)}=[a_{pq}^{(1,2;3,4)}],
\;
a_{ijkh}=a_{pq}^{(1,2;3,4)},
\end{equation}
where
$p=i+(j-1)I$ and $q=k+(h-1)K$.

The mode-$3$ product of a tensor $\AA\in\IR^{I\times J\times K}$ times a matrix
$U\in \IR^{H\times K}$ is defined by
$$
(\AA\times_3 U)(i,j,h)
=
\sum_{k=1}^K a_{ijk}u_{hk}.
$$
The tensor-matrix products in other modes are defined similarly.

The mode-$[3;3]$ contracted product of two third order tensors
$\AA\in\IR^{I_1\times J_1\times K}$ and $\BB\in\IR^{I_2\times J_2\times K}$
is defined by
$$
\langle\AA,\BB\rangle_{[3;3]}(i_1,j_1,i_2,j_2)
=
\sum_{k=1}^K a_{i_1i_2k}b_{i_2j_2k},
$$
where the first $3$ in $[3;3]$ indicates the third mode of $\AA$, and
the second $3$ refers to the third mode of $\BB$.
The definition can be generalized to different modes and multiple modes \cite{bk:tensor06}.

A useful property
is that given $\AA\in\IR^{I\times J\times K}$,
we let $\BB=\langle\AA,\AA\rangle_{[3;3]}$, and then have
\begin{equation}
\label{eqn:tensor_trace0}
\|\AA\|^2=\trace(B_{(1,2;3,4)}),
\end{equation}
where $B_{(1,2;3,4)}$ is the result of matricization of $\BB$.
See (\ref{eqn:matricize4}) for the definition.
With the tensor trace notation introduced in Section~\ref{sec:2dmethods},
we can write (\ref{eqn:tensor_trace0}) as
$
\|\AA\|^2=\trace(\langle\AA,\AA\rangle_{[3;3]}),
$
which is the high order generalization of
$
\|A\|=\trace(AA^T).
$

\section{Image-as-vector methods}
\label{sec:1dmethods}

We review the linear dimensionality reduction methods
which transform
$X=[x_1,x_2,\dots,x_n]\in \IR^{m\times n}$ into
$Y=[y_1,y_2,\dots,y_n]\in \IR^{d\times n}$ ($d<m$)
by a linear mapping $Y=U^TX$
in order to preserve some properties.
This matrix $U$ can be applied to project test data
for recognition tasks.
If the original numerical data items,
such as face images,
are not one-dimensional,
they are converted to column vectors
$x_1,x_2,\dots,x_n\in\IR^m$.
The mapping is called orthogonal,
if $U\in\IR^{m\times d}$ consists of orthonormal columns,
i.e., $U^TU=I_d$.

\subsection{Principal component analysis}
\label{sec:pca}

Principal Component Analysis (PCA) performs
an orthogonal mapping $Y=U^TX$ to maximize
the variance of the projected vectors.
More precisely, the objective function to maximize is
\begin{equation}
\label{eqn:pca_obj}
\sum_{i=1}^n \|y_i-\bar{y}\|^2
=
\|Y-\bar{y}e_n^T\|^2
=
\|Y-(\frac{1}{n}Ye_n)e_n^T\|^2
=
\|U^T X(I_n-\frac{1}{n}e_ne_n^T)\|^2
\end{equation}
subject to $U^TU=I_d$,
where $\bar{y}=\frac{1}{n}\sum_{j=1}^{n}y_j$ is the mean and
$e_n\in\IR^n$ is the column vector of ones.
Hence the maximizer of (\ref{eqn:pca_obj}) is
the left $d$ singular vectors of $X(I_n-\frac{1}{n}e_ne_n^T)$
corresponding to the largest $d$ singular values,
or equivalently the top $d$ eigenvectors
of $X(I_n-\frac{1}{n}e_ne_n^T)X^T$.
Note that $J_n:=I_n-\frac{1}{n}e_ne_n^T$ is a projection matrix
and therefore $J_n^2=J_n=J_n^T$.

In the methods described subsequently in this appendix,
it is common to pre-process the training data matrix $X$ by PCA,
in order to avoid a singular matrix in the eigenvalue computation.
By abuse of notation, we also denote by $X$
the resulting matrix preprocessed by PCA.

\subsection{Affinity graph}
\label{sec:affinity_graph}

Several dimensionality reduction methods
utilize an affinity graph $G=(V,E)$,
where $V=\{1,\dots,n\}$ consists of indices of data items.
Data item $j$ is deemed related to
data item $i$ if $(i,j)\in E$.
If the input data are unsupervised,
one may construct a $\kNN$ graph or an $\epsilon$-graph
for an affinity graph.
The graph is made undirected if symmetry is desired.
In supervised learning,
each data item is associated with a class label.
For example, in face recognition,
each training image is of subject.
A label graph $G=(V,E)$ is defined by that
$(i,j)\in E$ if data items $i$ and $j$
have the same label, i.e., in the same class.
It has been observed that
supervised graphs often outperform
their unsupervised peers in the recognition tasks.

Each edge $(i,j)\in E$ is associated with a weight $w_{ij}$
as the measure of influence
between the two neighboring points $i$ and $j$.
A popular choice is the {\em Gaussian weights}:
\begin{equation}
\label{eqn:gaussian_weights}
w_{ij} =
\left\{
\begin{array}{ll}
e^{-\|x_i-x_j\|^2/t} & \mbox{if }(i,j)\in E,\\
0 & \mbox{otherwise},
\end{array}
\right.
\end{equation}
where $t>0$ is some constant.
Alternatively, we can use the {\em binary weights}
from driving $t\rightarrow 0$.
It is common to employ an undirected affinity graph
when the Gaussian weights or the induced binary weights are used.

Another weighting scheme, proposed in
the {\em Locally Linear Embedding} (LLE) \cite{rs:lle00},
is from minimizing the function
\begin{equation}
\label{eqn:llew_prog}
\left\{
    \begin{array}{ll} \displaystyle
    \mini_{w_{ij}} & \displaystyle \sum_{i=1}^n \|x_i-\sum_{j=1}^n w_{ij}x_j\|^2 \\
    \textup{subject to} & w_{ij}=0\mbox{ for }(i,j)\notin E, \\
                        & \sum_{j=1}^n w_{ij}=1\mbox{ for all }i.
    \end{array}
\right.
\end{equation}
subject to $w_{ij}=0$ if $(i,j)\notin E$, and
$\sum_{j=1}^n w_{ij}=1$ for $i=1,\dots,n$.
The minimizer of (\ref{eqn:llew_prog}) is obtained
from solving $n$ symmetric linear systems.
See \cite{rs:lle00} for more information.
Note that the resulting weights are usually asymmetric
and can be negative.
The discussion on the affinity graph and the weighting scheme
applies not only to this appendix
but also to Sections~\ref{sec:2dmethods} and \ref{sec:trepulsion}.

\subsection{Locality preserving projection}
\label{sec:lpp}

Consider the objective function to minimize:
$
\frac{1}{2}\sum_{i,j=1}^n w_{ij}\|y_i-y_j\|^2,
$
where $w_{ij}$'s are non-negative weights,
e.g., the Gaussian weights (\ref{eqn:gaussian_weights}).
In what follows
we need the weight matrix $W=[w_{ij}]\in R^{n\times n}$ being symmetric.
Let $D\in\IR^{n\times n}$ be the diagonal matrix
formed by $d_{ii}=\sum_{j=1}^n w_{ij}$, and
denote the graph Laplacian by $L=D-W$.
After some algebra, we have
\begin{equation}
\label{eqn:le_obj}
\frac{1}{2}\sum_{i,j=1}^n w_{ij}\|y_i-y_j\|^2
=
\trace(YLY^T).
\end{equation}

{\em Laplacian eigenmaps} \cite{bn:eigenmaps03} is
a nonlinear dimensionality reduction method
that minimizes (\ref{eqn:le_obj})
subject to $YDY^T=I_d$ and $YDe_n=0$.
{\em Locality Preserving Projection} (LPP) \cite{hn:lpp03,hyunz:laplacian05}
also minimizes (\ref{eqn:le_obj}),
but imposes the linear projection constraint $Y=U^TX$
and solves
\begin{equation}
\label{eqn:lpp_prog}
\left\{
    \begin{array}{ll} \displaystyle
    \mini_{U} & \displaystyle \trace(U^TXLX^TU) \\
    \textup{subject to}   & U^TXDX^TU=I_d.
    \end{array}
\right.
\end{equation}
This is equivalent to solving the generalized eigenvalue problem
\begin{equation}
\label{eqn:lpp_sol}
X L X^T u_i = \lambda_i X D X^T u_i.
\end{equation}
The minimizer of the program (\ref{eqn:lpp_prog}) is
$$
U=[u_1,\dots,u_d]\in\IR^{n\times d},
$$
where $u_1,\dots,u_d$ are the eigenvectors corresponding to
the smallest generalized eigenvalues of (\ref{eqn:lpp_sol}).
Note that the program (\ref{eqn:lpp_prog}) can be regarded as a workaround
to minimize $\trace(U^TXLX^TU)$ and maximize $\trace(U^TXDX^TU)$
concurrently.

Another option is to replace the constraint $U^TXDX^TU=I_d$
in (\ref{eqn:lpp_prog}) by simply the orthogonal mapping,
i.e., $U^TU=I_d$, and then
the minimizer $U$ of (\ref{eqn:le_obj})
consists of the $d$ eigenvectors of $XLX^T$,
corresponding to the $d$ smallest eigenvalues.
We call this method
the {\em Orthogonal Locality Preserving Projection}
(OLPP) \cite{ks:onpp07,ks:repulsion09}.
In practice, we use the supervised label graph and
the Gaussian weights (\ref{eqn:gaussian_weights})
for both LPP and OLPP.

\subsection{Neighborhood preserving projection}
\label{sec:npp}

The nonlinear dimensionality reduction method
LLE \cite{rs:lle00} minimizes
\begin{equation}
\label{eqn:lle_obj}
\sum_{i=1}^n \|y_i - \sum_{j=1}^n w_{ij}y_j\|^2
=
\|Y-YW^T\|^2
\end{equation}
subject to $YY^T=I_d$ and $Ye_n=0$.

One can impose the linear projection $Y=U^TX$ and solve
\begin{equation}
\label{eqn:npp_prog}
\left\{
    \begin{array}{ll} \displaystyle
    \mini_{U} & \displaystyle \|U^TX(I_n-W)^T\|^2 \\
    \textup{subject to}   & U^TXX^TU=I_d.
    \end{array}
\right.
\end{equation}
We call the resulting method {\em Neighborhood Preserving Projection} (NPP)
\cite{hcyz:npe05}.
The minimizer of (\ref{eqn:npp_prog}) consists of the $d$ eigenvectors
corresponding to the $d$ smallest generalized eigenvalues of
$$
X(I_n-W)^T(I_n-W)X^Tu_i=\lambda_i XX^Tu_i.
$$

A related method, {\em Orthogonal Neighborhood Preserving Projections} (ONPP)
\cite{ks:onpp05,ks:onpp07}, solves
\begin{equation}
\label{eqn:onpp_prog}
\left\{
    \begin{array}{ll} \displaystyle
    \mini_{U} & \displaystyle \|U^TX(I_n-W)^T\|^2 \\
    \textup{subject to}   & U^TU=I_d,
    \end{array}
\right.
\end{equation}
which replaces $U^TXX^TU=I_d$ in (\ref{eqn:npp_prog}) by
the $U^TU=I_d$.
The solution is formed by the $d$ left singular vectors of $X(I_n-W)^T$
corresponding to the $d$ smallest singular values, or equivalently
the bottom $d$ eigenvectors of $X(I_n-W)^T(I_n-W)X^T$.
Note that both NPP and ONPP do not need weights being symmetric.
Therefore, it is common to adopt
the weighting scheme (\ref{eqn:llew_prog}) of LLE.

It is interesting to note that
$(I_n-W)^T(I_n-W)$ in NPP and ONPP plays the role of the
Laplacian matrix $L$ in LPP and OLPP \cite{kcs:trace10}.
This observation is important to the repulsion techniques
in \cite{ks:repulsion09} and in this paper.

\subsection{Linear discriminant analysis}
\label{sec:lda}

The method of {\em Linear Discriminant Analysis} (LDA) \cite{bhk:fisher97,fisher:lda36}
can be formulated as follows.
Suppose we are given high dimensional data
$X=[x_1,x_2,\dots,x_n]\in\IR^{m\times n}$ with
each data sample $x_i$ associated with a class label $c(i)$.
Let
$$
\CC_j=\{i:c(i)=j\}
$$
be the index set of all data items in class $j$,
and $n_j=|\CC_j|$ be the size of class $j$.
The mean of each class $j$ is denoted by
$\bar{x}_j=\frac{1}{n_j}\sum_{i\in\CC_j}x_i$,
and the global mean is $\bar{x}=\frac{1}{n}\sum_{i=1}^n x_i$.
The within-scatter matrix of $X$ is defined by
\begin{equation}
\label{eqn:Sw}
S_w=\sum_j\sum_{i\in\CC_j}(x_i-\bar{x}_j)(x_i-\bar{x}_j)^T,
\end{equation}
and the between-scatter matrix of $X$ is
\begin{equation}
\label{eqn:Sb}
S_b=\sum_j n_j(\bar{x}-\bar{x}_j)(\bar{x}-\bar{x}_j)^T.
\end{equation}
Since the lower dimensional data are obtained by
a linear mapping $Y=U^TX\in\IR^{d\times n}$,
the between-scatter matrix
and within-scatter matrix of $Y$ are
$U^TS_wU$ and $U^TS_bU$, respectively.

We would like to maximize $\trace(U^TS_bU)$ and minimize $\trace(U^TS_wU)$
in some way.
It is common to solve
\begin{equation}
\label{eqn:lda_prog}
\left\{
    \begin{array}{ll} \displaystyle
    \maxi_{U} & \displaystyle \trace(U^TS_bU) \\
    \textup{subject to}   & U^TS_wU=I_d,
    \end{array}
\right.
\end{equation}
The maximizer of (\ref{eqn:lda_prog}) is $U=[u_1,\dots,u_d]$,
the $d$ largest generalized eigenvalues of
$$
S_b u_i = \lambda_i S_w u_i.
$$
Note that LPP can be regarded a generalization of LDA.
See \cite{hyunz:laplacian05} for a discussion.

In comparison, 
the projections of LPP, OLPP, NPP, and ONPP
can be performed in either supervised or unsupervised mode,
whereas
the PCA projection is unsupervised and
the LDA projection is supervised.

\end{document}